\documentclass{article}
\usepackage[square,numbers]{natbib}
\usepackage[main,final]{neurips_2026}
\usepackage{multirow}

\usepackage{microtype}
\usepackage{graphicx}
\usepackage{subcaption}
\usepackage{booktabs} 
\usepackage{url}

\usepackage[compact]{titlesec}

\usepackage{hyperref}

\usepackage{inconsolata}

\usepackage{amsmath}
\usepackage{amssymb}
\usepackage{mathtools}
\usepackage{amsthm}

\usepackage[utf8]{inputenc} 
\usepackage[T1]{fontenc}    
\usepackage{hyperref}       
\usepackage{url}            
\usepackage{booktabs}       
\usepackage{amsfonts}       
\usepackage{nicefrac}       
\usepackage{microtype}      

\usepackage[capitalize,noabbrev]{cleveref}

\theoremstyle{plain}

\theoremstyle{definition}

\theoremstyle{remark}

\usepackage[dvipsnames]{xcolor}
\usepackage[textsize=tiny]{todonotes}

\usepackage[skip=2pt]{caption}
\usepackage{paralist}

\newcommand{\para}[1]{\smallskip\noindent\textbf{#1}~}

\usepackage{enumitem}

\author{%
  Eve Fleisig $^{1, 2\thanks{Work conducted while visiting Università Bocconi. Corresponding author: \texttt{efleisig@berkeley.edu}}}$ \: Joachim Baumann$^{2, 3}$ \: Dylan Hadfield-Menell$^4$ \: Dirk Hovy$^2$\\
  $^1$UC Berkeley \: \:
  $^2$Universit\`{a} Bocconi  \: \:
  $^3$Stanford University \: \:
  $^4$MIT
}

\title{Leaderboard Hacking: Preference-Based Model Evaluations are Vulnerable to Manipulation}

\begin{document}


\maketitle





\begin{abstract}
  Arena-style rankings of language models are a widely used evaluation framework in online leaderboards, research papers, and model evaluation in production. These rankings aim to  realistically assess  model performance using pairwise comparisons of anonymous models on user prompts, aggregating  votes via the Bradley-Terry model to produce the final ranking. However, we find that these leaderboards are highly vulnerable to several manipulation strategies that unscrupulous providers could use to boost a  model's rank: \textit{strategic voting} (individual votes that boost a model's rank, including votes on irrelevant models), \textit{strategic prompting} (choosing prompts that favor a model), and \textit{strategic nomination} (boosting the rank of a model by inserting irrelevant models). We demonstrate these effects on public leaderboard data, analyze the circumstances that cause these vulnerabilities, and describe approaches to safeguard against each attack. Notably, these issues hold for any  preference-based evaluation using the Bradley-Terry model that allows open prompting, voting, or model submission. These manipulations boost the rank of nearly every tested model; each attack can boost a model by up to 15-20 places in a ranking and, combined, they can boost a model by up to 45 places.
\end{abstract}

\section{Introduction}
How can we compare the performance of large language models (LLMs)? To address concerns that static benchmarks do not evaluate real-world usage, arena-style evaluations offer a more flexible alternative: users post a prompt of their choice and indicate their preference between two anonymous models' responses. These pairwise preferences are then aggregated and transformed into rankings on prominent online leaderboards, such as Artificial Analysis (\url{artificialanalysis.ai}), DesignArena (\url{designarena.ai}), and LMArena (\url{lmarena.ai}). However, as these preference-based rankings gain influence---they affect users' decisions on which model to use, company publicity, funding decisions, and ultimately model developers' revenue---incentives increase to win these competitions even by unscrupulous means. We consider: how much, and how easily,  could a bad actor fraudulently boost a model's rank?

Most preference-based evaluations transform a set of pairwise preferences into a unified ranking using the Bradley-Terry model, which implicitly reduces to the Borda count \citep{siththaranjandistributional}, a type of positional voting with known vulnerabilities. Because of this, we hypothesized that known weaknesses of the Borda count also hold for LLM rankings that use Bradley-Terry. We test this hypothesis and indeed find three simple strategies that allow fraudulent actors to boost the rankings of an arbitrary \textit{target model} $A$. 

In \textit{strategic voting}, users change their votes on competitor models to boost $A$'s rank. Because the strength of $A$'s competitors affects $A$'s ranking, even strategic votes on matches that do not involve $A$ can influence the ranking in its favor. In \textit{strategic nomination}, users insert copies of irrelevant models to improve the rank of $A$. By inserting copies of models against which $A$ does particularly well relative to its competitors, this method boosts the target model's ranking compared to other models. Finally, in \textit{strategic prompting}, users provide prompts that favor $A$, e.g., prompts in a language on which $A$ excels. Because win rates vary widely when splitting the preference data into subsets, $A$'s rank can be boosted by prompting in any definable subcategory where $A$ does well.
As a corollary, we find that it is surprisingly easy to identify which models are presented, even though anonymity is a cornerstone of these evaluations.

The methods we find for manipulating model rankings are easy to implement, and can be combined to magnify the boosting effect. We describe each method in detail and demonstrate their effects on publicly available data from LMArena. Our contributions are (1) we demonstrate the effect of strategic voting, nomination, and prompting on preference rankings and link to the social choice literature (equivalence to Borda count); (2) we introduce an efficient strategic voting heuristic (that does not require inefficient Elo recomputations); and (3) we propose defenses against each attack.

\section{Background and Related Work}
\label{sec:background}

\subsection{Preference-based evaluation and its vulnerabilities}
The Bradley-Terry model (BT) estimates the relative strength of items based on individual preferences between pairs of items. Let $(A, B)$  denote a comparison between candidates $A$ and $B$, and let $A \succ B$ denote that $A$ is preferred to $B$: BT assumes that human preferences obey $P_{BT} (A \succ B) = \frac{e^{\xi_A}}{e^{\xi_A} + e^{\xi_B}}$. 

The coefficients $\xi$ serve as scores for the competitors (i.e., models on a leaderboard). \citet{boubdir2024} and \citet{bai2022traininghelpfulharmlessassistant} discuss using preference-based evaluations for language models. \citet{chiang2024} build on this to rank LLMs from user preferences; they estimate BT scores efficiently using maximum likelihood estimation on logistic regression, which we follow in this work.\footnote{On LMArena, these scores are sometimes referred to as Elo scores. The Elo model is a linear scaling of the Bradley-Terry model, but more recent comparisons are weighted more heavily when fitting the model (unlike Bradley-Terry, which weights comparisons equally regardless of when they occurred). However, the current LMArena implementation uses Bradley-Terry scaled to Elo's scoring range (models are assumed to be static), not Elo; see \url{github.com/lmarena/arena-rank}.}

\citet{singh2025leaderboardillusion} raise issues regarding the reliability of LMArena's rankings due to pre-release testing that results in uneven data access for model providers, uneven sampling, and silent model deprecation; they argue that these differences benefit some model providers over others.\footnote{The extent of these concerns is debated; see \url{lmarena.ai/blog/our-response}} In contrast, the evaluation vulnerabilities we discuss here are not limited to any particular leaderboard. Leaderboards designed impartially and in good faith remain vulnerable to the issues we present so long as they rely on the Bradley-Terry model.

Some previous work has explored methods of adversarial attacks on preference-based leaderboards. \citet{zhao2025challenges} and \citet{huang2025exploring} consider strategic voting attacks based on identifying the target model and then voting strategically on examples that involve that model. \citet{min2025improving} expand this to consider strategic voting on a stream of random votes on pairs $(B,C)$ by considering what a target model's Elo score would be after voting in favor of $B$ or $C$. In concurrent work, \citet{hays2026strategiccandidacygenerativeai} explore strategic candidacy restricted to cloning, in which model providers can benefit from submitting copies of the target model itself. \citet{liu2025reevaluatingopenendedevaluationlarge} pinpoint weaknesses in permitting open-ended prompting for LLM evaluation, since models could be trained to specialize in the skills that are overrepresented in the prompts. 
\citet{pmlr-v235-zhang24u} study multi-task benchmarks through social choice theory and show that adding irrelevant models can change aggregate rankings in ordinal benchmarks.
\citet{chen2026leaderboard} model leaderboard competition as a game in which developers invest in benchmark-specific post-training, and analyze tune-before-test as an intervention on these incentives.
Relative to this work, our contributions are as follows:

\begin{itemize}[nosep,leftmargin=11pt]
\item We introduce two new attacks, strategic nomination and strategic prompting, based on the ability to submit arbitrary prompts or models on these arenas.
\item We introduce an efficient strategic voting heuristic, with no inefficient Elo recomputation per vote.
\item We explain why strategic voting, nomination, and prompting are effective by linking to the social choice literature (equivalence to Borda count).
\item We show that the new attacks we discuss result in large boosts to many models' rankings.
\item We propose defenses against each attack.
\end{itemize}

\subsection{Parallels to voting vulnerabilities} \citet{siththaranjandistributional} found that evaluations using the Bradley-Terry model implicitly optimize for the Borda count, a positional voting rule in which candidates in an election are scored based on how many candidates are ranked below them. In the traditional Borda count formulation, for an election with $n$ candidates, each voter ranks every candidate in order of preferences. The first-place candidate for that voter receives $n$ points; second place $n-1$, etc. These points are summed across voters, and candidates are ranked in order of points. This is equivalent to ranking candidates in order of the average probability (over voters) that the candidate is preferred to other candidates in the election. \citet{siththaranjandistributional} find that any utility function that optimizes for the BT objective implicitly aggregates individual voters' utilities according to the Borda count.
However, in the social choice literature, the Borda count is known to be vulnerable to several kinds of strategic manipulation \cite{brandt2016handbook}.
Thus, we hypothesize that \textbf{preference-based leaderboards that implicitly optimize for the Borda count inherit these weaknesses.} 
In particular, we hypothesize that one can manipulate these leaderboards to raise a model $A$'s rank relative to a model $B$ using:

\textbf{Strategic voting.} Suppose that $A$ and $B$ are the top two candidates for an election, and supporters of a candidate $A$ wish for it to win over candidate $B$. Under the Borda voting rule, they can use \textit{strategic voting}, in which supporters vote that they prefer a third candidate $C \succ B$ in order for $A$ to win (even though they truly believe $B \succ C$). In LLM leaderboards, users can raise $A$'s rank over $B$ by strategically voting on comparisons with unrelated models. They can also abstain from votes on pairings that do not involve $A$ even though they do truly have a preference, in order to increase the fraction of their votes involving $A$. In practice, strategic voting requires deanonymizing models in order to vote in favor of $A$.

\textbf{Strategic nomination.} Fraudulent actors can also engage in \textit{strategic nomination}, by introducing a candidate $C$ such that $A \succ C \succ B$. For voters who initially preferred $A \succ B$, their new ranking is $A \succ C \succ B$, widening $A$'s lead; for voters who initially preferred $B \succ A$, their new ranking is $B \succ A \succ C$, boosting $A$ and $B$ equally. Repeating this process with many candidates $C$ can substantially lower B's Borda score relative to A. In elections, this is most often discussed in the context of cloning, in which $C$ is very similar to $A$ but slightly worse \citep{brandt2016handbook}. Users can add clones $A'$ of a model $A$, or copies of unrelated models $C$, in ways that raise model $A$'s rank relative to $B$. If the voters simply want one of the models that resembles $A$ to win, we may also relax the restriction that $A'$ be similar to $A$: this means that it is to a faction's advantage to nominate many similar candidates. More broadly, inserting copies of any model that usually wins over $B$ but loses to $A$ will improve $A$'s rank.

\textbf{Strategic prompting.} Finally, LLMs offer a domain-specific way to raise model $A$'s rank by exploiting the fact that Borda count aggregates over all votes evenly. If voters know that there is a particular definable area in which $A$ excels relative to other models, they can simply choose prompts in that area to explicitly favor $A$ over $B$. This takes advantage of high variance in model win rates between different areas, defined broadly: e.g., different skills (coding vs. writing), or different languages (English vs. Korean vs. Amharic). For example, if model $A$ does very well in Amharic but other models cannot process it, a voter can simply always prompt the model in Amharic.  
    
We can consider strategic prompting as a type of strategic voting specific to language models. Live strategic voting requires deanonymizing models in order to know which model to vote for. Instead, strategic prompting uses subcategory performance as a proxy for model identification: if model $A$ typically performs well on prompts in a subcategory, then in expectation, prompting in that subcategory and always voting for the better model is a strategy that makes it more likely to vote in favor of $A$.

For each of the three attacks, we consider how feasible, likely, and preventable the attack is along three axes:
\begin{enumerate}[nosep]
    \item How many strategic votes, prompts, or inserted models are needed to shift a ranking?
    \item How easy is it to gather the information required (e.g., identify anonymized models, create a suitable model clone, or determine which way to vote strategically)?
    \item How can leaderboard developers secure their evaluations against these attacks?
\end{enumerate}


\section{Methods}\label{sec:methods}
To examine the effects of leaderboard manipulation on real data, we conduct our experiments on a public leaderboard dataset of over 100,000 human preferences from LMArena.\footnote{The data is at \label{data}\url{huggingface.co/datasets/lmarena-ai/arena-human-preference-100k} under a CC-BY-4.0 license.}

\subsection{Strategic Voting}
We consider two strategic voting attacks: \emph{greedy best-vote}---a computationally expensive method explored in \cite{min2025improving}---and \emph{heuristic best-vote}---an efficient heuristic strategy we introduce here. Both strategies leverage Borda count and Bradley-Terry's dependence on irrelevant alternatives: votes on model pairs that do not involve the model $A$ can nonetheless boost $A$'s rank.

\para{Greedy Best-Vote} We consider an attack in which the manipulator votes strategically on a stream of new comparisons between random pairs of models (many of which do not involve $A$). We see that votes on unrelated models do shift $A$'s Elo score, since they shift the scores of $A$'s competitors. For this attack, we run a greedy best-strategic-vote algorithm: given a set of models $S$ that includes a target model $A$ to move up in the rankings, and a stream of $n$ new comparisons to perform between any pair of models $(B, C) \in S$, check whether voting for model $B$ or model $C$ results in the best improvement for model $A$.\footnote{We initially experimented with allowing ties, but found that voting for one winner was almost always better at shifting model ranks; thus, we restrict the options considered by this strategy to voting for one winner, reducing computational cost.}  Then, calculate the improvement for $A$ under $n$ strategic votes. For these experiments, we set $n$ to be at most 10K votes ($\sim$10\% of votes in the dataset). 

\citet{min2025improving} explore a greedy best-vote strategy in prior work (for each new preference pair, vote for the choice that best improves the target model's Elo). However, the greedy best-vote strategy is computationally expensive because it recalculates Elo scores after every vote.

\para{Heuristic Best-Vote} We introduce a simple efficient heuristic for strategic voting that avoids recomputing Elo scores. Our initial experiments with greedy best-vote found that flipping the winner of a preference is typically optimal for boosting a model's rank. Based on this, we implement a simple voting rule for a pair $(B, C)$, where the voter's true preference is $B \succ C$: If $A \in \{B, C\}$, vote for $A$; if not, vote for the worse model $C$.

\para{Identifying models in order to vote strategically.} In practice, strategic voting requires the manipulator to \textit{know} which pair of models they are voting on. Extensive prior work focuses on detecting LLM-generated text, typically by training classifiers to distinguish human-written from model-generated content or to attribute text to a specific model~\citep{uchendu-etal-2020-authorship,verma-etal-2024-ghostbuster,sadasivan2025can}.
Our setting is much simpler, since the evaluator has full control over the  prompt. The simplest identification strategy is to prompt the model to reveal its own identity. However, LMArena filters out responses that include the model name \citep{huang2025exploring}. We therefore consider an approach that is undetectable with output filtering.
For open-weight models, we use a likelihood-based approach: given a prompt-response pair randomly sampled from the LMArena data,\textsuperscript{\ref{data}} we compute the average token-level log-likelihood of the response for a set of candidate models by measuring the probability assigned to each next answer token conditioned on all previous tokens. Specifically, we run each candidate open-weight model over the observed response in teacher-forcing mode, after applying that model's own chat template, and compute the average token-level log-probability that the candidate assigns to that exact response. We then identify models based on the highest average log-probability. For proprietary API-based models, where token-level log-probabilities are unavailable, we use an even simpler approach. We generate a single response for a candidate model and score similarity to the observed response using character-level sequence similarity.

\subsection{Strategic Nomination}
In strategic nomination, the objective is to find a model $C$ for which adding copies of $C$ boosts $A$'s ranking. To do so, we (1) identify the model $C$ that best benefits model $A$, and (2) add new copies $C'$ whose performance is assumed to be identical to $C$ (thus, its outcomes in battles with existing models are likely the same). Computing the exact solution for which model to add is cheap, since there are only 55 candidate models in our dataset (and about 300 on the current live version of LMArena). For each target model $A$, we find the model $C$ for which adding $n$ copies of model $C$ and its battles results in the best rank improvement for $A$; then, we add $n$ copies of model $C$ and recompute model $A$'s rank. For these experiments, we set $n$ at $\leq$10 copies.

\subsection{Strategic Prompting}
We consider how to manipulate votes simply by restricting prompts to a subset on which model $A$ outperforms other models. As proof-of-concept, we use language to define our prompt subsets, for several reasons: the language of the prompt is readily available categorical data in the LMArena dataset, it is easy for a manipulator to restrict their prompts to that subset, and it is easy to create new prompts within any given subset by simply prompting in that language. However, we note that any easy-to-define split under which $A$ performs well relative to other models would work here (e.g., performance on specialized domains); see Appendix \ref{sec:prompting-other-subcats} for an example.

For this attack, we first split the data into prompt subsets (in our case, different languages). We identify the language in which model $A$ is ranked highest relative to the other models\footnote{We restrict to languages with at least 20 battles.} and insert $n$ random battles with known results in that language (i.e., copies of existing battles). For these experiments, we set $n$ to be at most 10K votes (approximately 10\% of votes in the dataset). As a side effect, we note that strategic prompting also poisons later evaluations that use the prompt set: e.g., if prompts from an arena are later used for offline evaluation of new models, but strategic prompting in favor of a model $A$ was used on evaluations on the arena, then those later evaluations will also favor models similar to $A$.

\section{Results}
\subsection{Strategic Voting}
\begin{figure*}[!ht]
  \begin{center}
    \centerline{\includegraphics[width=0.66\textwidth]{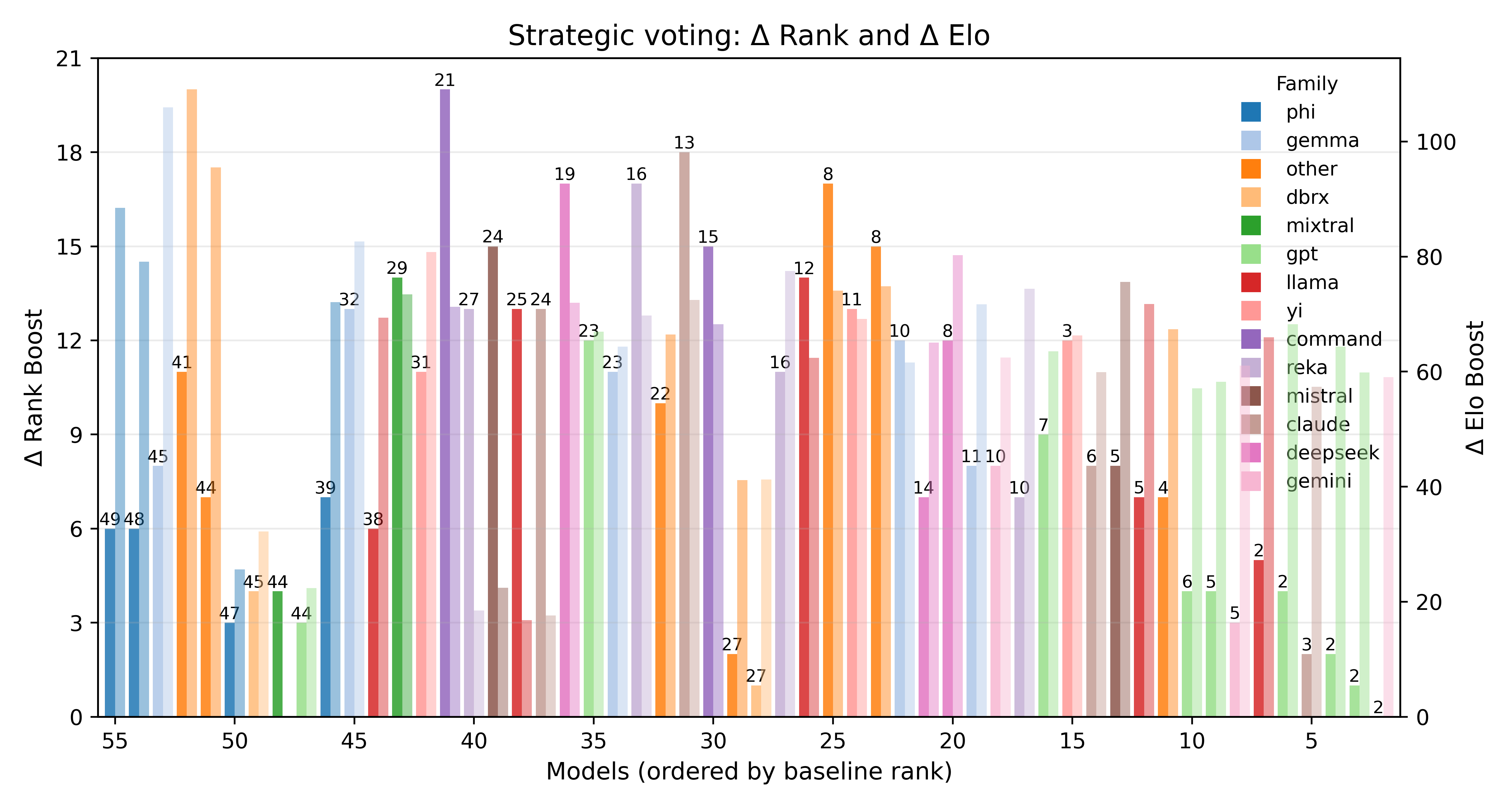}}
    

    \centerline{\includegraphics[width=0.66\textwidth]{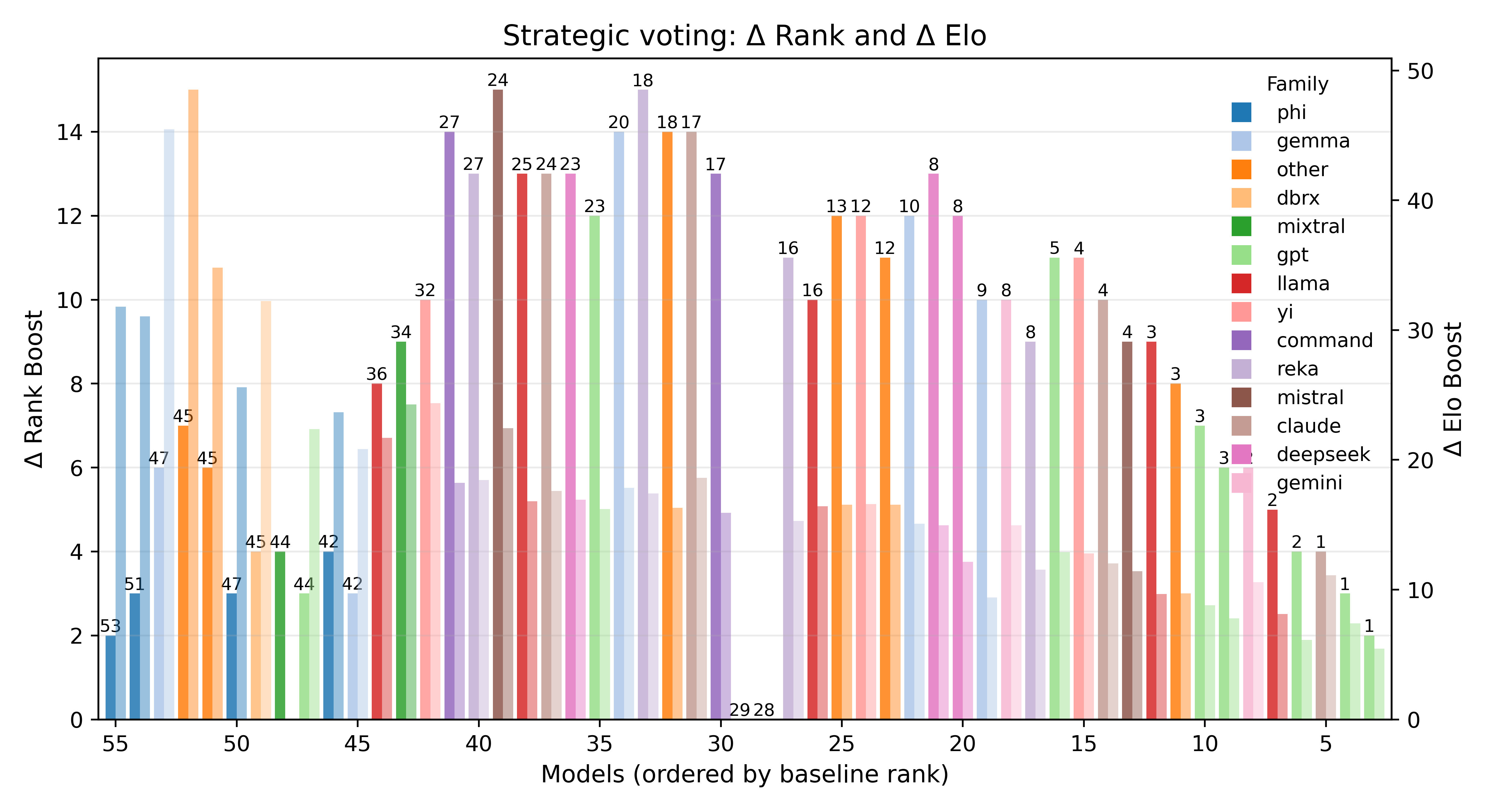}}
    \caption{
      \textit{Top:} Highest model rank (dark bars) and Elo (light bars) achieved under \textbf{greedy} strategic voting, capped at 10K votes ($\sim10\%$ of total votes). Numbers above bars indicate the best absolute rank that a model achieves. Models are boosted up to 20 places in the rankings; 23 models (42\% of the dataset) move over 10 places up, and eight models move into the top 5. \textit{Bottom:} Highest model rank \& Elo achieved under  \textbf{heuristic} strategic voting, under same conditions. Models are boosted up to 15 places in the rankings; 21 models (38\% of the dataset) move over 10 places up, and 15 models move into the top 5.
    }
    \label{fig:vote_rank_elo_change_heuristic}
  \end{center}
\end{figure*}


\para{Strategic Voting (Greedy Strategy)} Adding strategic votes substantially boosts target models' ranks (Figure \ref{fig:vote_rank_elo_change_heuristic}, top). We find that \textbf{models are boosted up to 20 places} in the ranking using this method, and that every tested model below 2nd place can move up at least one place. Boosting a model higher in the rankings requires more votes (Figure \ref{fig:votes_needed_for_mvt}), and models close to the very top of the leaderboard (2nd-4th place) are harder to move than models further down, in part because they have participated in more battles. However, 23 models (42\% of the dataset) move over ten places upwards using this method, and rankings near the top of the leaderboard are still swayed relatively easily: 11 models can move into the top 10 places, and eight models can move into the top 5 places.

\para{Strategic Voting (Heuristic)} The strategic voting heuristic also substantially boosts target models' ranks (Figure \ref{fig:vote_rank_elo_change_heuristic}, bottom). Models are \textbf{boosted up to 15 places}, and 21 models move over ten places upward. This method is particularly effective for models high up in the ranking, moving fifteen models into the top 5 places. This suggests that, while the greedy voting strategy results in the largest boost for the most models, the heuristic strategy results in larger boosts for models near the top of the rankings, at a far lower computational cost.

Figure \ref{fig:votes_needed_for_mvt} (Appendix \ref{app:model-family}) compares the number of votes needed under each strategy to boost a model's rank. Small improvements to rank (2-3 places) are relatively easy to achieve, requiring under 2\% of total votes (2K out of 100K total) to be manipulated and achievable by a relatively large subset of models. Larger improvements (15+ places) are achievable by a smaller subset of models with more manipulated votes. We find that while the maximum rank boost achieved across models is higher for the greedy strategy, more models are able to achieve smaller boosts under the heuristic strategy (and typically with fewer votes). We also find that, in practice, \textbf{the identification methods for open-weight and proprietary models suffice for heuristic strategic voting} (Appendix~\ref{app:live_model_identification}).
Likelihood-based attribution identifies open-weight models with high precision ($>80\%$), and similarity-based scoring for API-access models performs above chance ($>$40-50\%).

\subsection{Strategic Nomination}
\begin{figure*}[!ht]
  \vskip 0.2in
  \begin{center}
    \centerline{\includegraphics[width=0.66\textwidth]{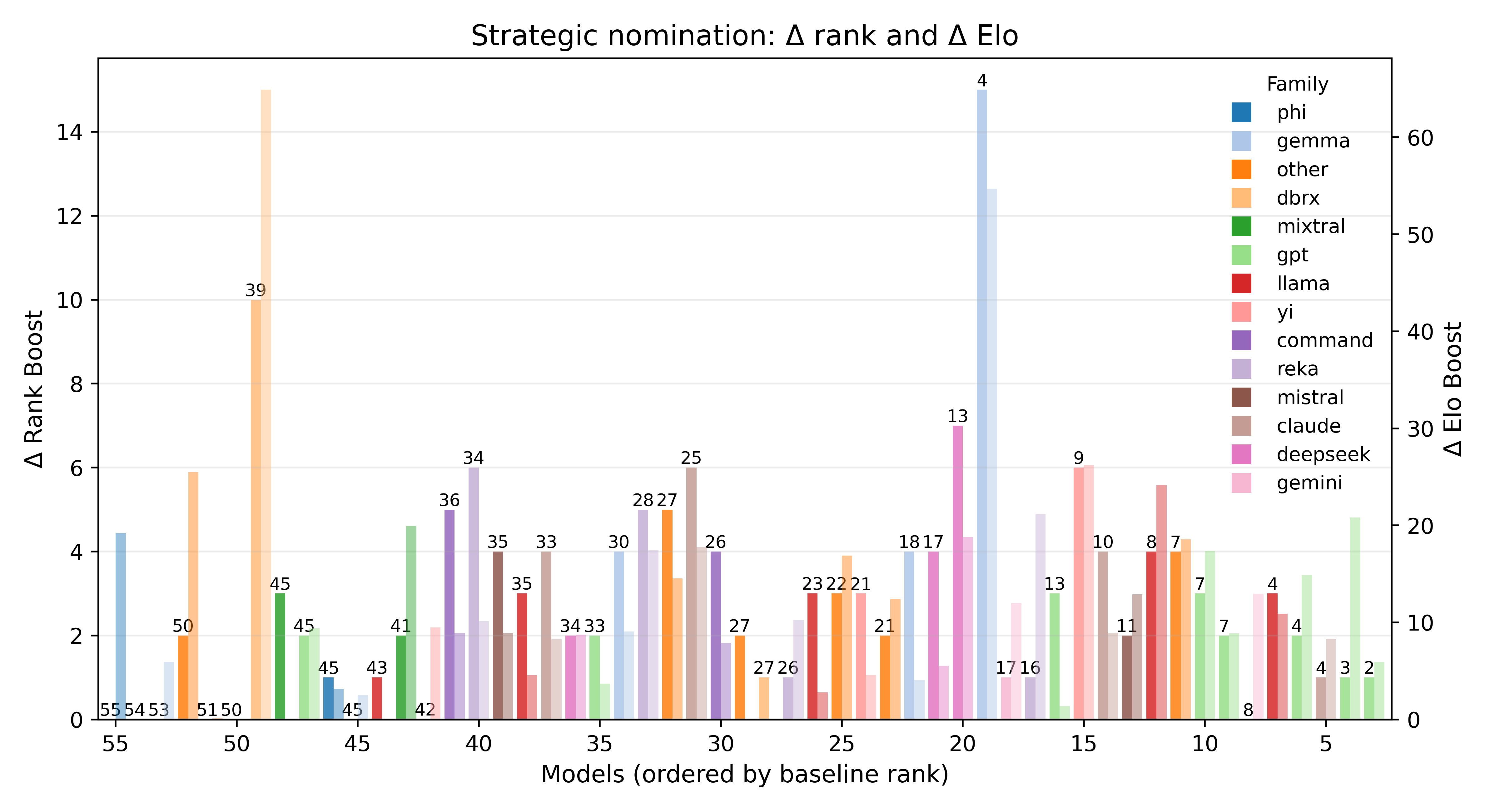}} 
    \caption{
      Change in model rank and Elo achieved when adding strategic nominations (10 copies of the most favorable model). Change in rank is the left-hand bar of each double-bar; change in Elo is the right-hand (lighter) bar. Numbers above bars indicate the best absolute rank achieved by the model (1=best). For example, the model originally in 19th place (light blue) jumps 15 places to 4th place.
    }
    \label{fig:nom_rank_elo_change}
  \end{center}
\end{figure*}

Strategic nomination \textbf{boosts model ranks by up to 15 places} (Figure \ref{fig:nom_rank_elo_change}), though  boosts are typically more moderate (1-7 places). Curiously, the effect of strategic nominations varies widely by  model; for example, the 19th place model jumps to 4th place, whereas other models with similar rankings display more moderate effects. We examine these factors in more detail in Section \ref{sec:regressions}.

\para{Nomination effects in existing leaderboards.}
We examine whether these effects might already affect existing leaderboards by examining a counterfactual (Figures \ref{fig:family-removal-noms-up} and \ref{fig:family-removal-noms-down}, Appendix \ref{app:model-family}): what would a model's rank be if a particular family of models were included or excluded? A more stable ranking would be immune to irrelevant alternatives, i.e., in a set of models $S$, $A$'s rank relative to $S \setminus B$ should not depend on whether $B$ is included. For each model $A$, we find the family $B$ of models whose removal increases $A$'s rank the most, and the family that decreases $A$'s rank the most. We also calculate $A$'s rank in the full (original) dataset, skipping members of family $B$ for its ranking. We compare each model $A$'s rank among $S \setminus B$ when family $B$ is excluded from the Elo calculation, versus its rank among $S \setminus B$ when family $B$ is included. We thus measure whether the ranking among a set of models is affected by the inclusion of unrelated models.

We find that \textbf{including a model family $B$ can shift a model's rank among $S \setminus B$ up to 1-4 places in a desired direction (up or down)}. Claude models tend to have the strongest effect on model ranks, causing the most movement downwards for 12 models and upwards for 10: this is likely because the strongest Claude model participates in more battles than any other model (14,000, compared to 10,000 for the model with the second-most battles). Several models are also boosted by the inclusion of models from their own family: the inclusion of other Gemma models moves \texttt{gemma-2-9b-it-simpo} up two places, and the inclusion of other Claude models moves Claude 3 Sonnet up two places as well. Since each model family covers many battles, we compare with removing the same number of random battles and find that random removals can account for some but not all of this variation; boosts from the presence of the ChatGPT, LLaMA, and Command-R families are beyond what random removals can achieve. Furthermore, \textbf{42\% of models among $S \setminus B$ are boosted up one or more places in the ranking due to a single model}. For each removed model, we compare with the movement achievable by removing an equivalent number of random battles over 5 trials and find that for 38\% of models, a model's inclusion or removal results in a higher boost than random removal can achieve. Claude 3.5 Sonnet, the model involved in the most battles, has the strongest effects: it moves six models up 1-3 ranks when it is included compared to when it is excluded.

\subsection{Strategic Prompting}
\begin{figure*}[!ht]
  \begin{center}
    \centerline{\includegraphics[width=0.69\textwidth]{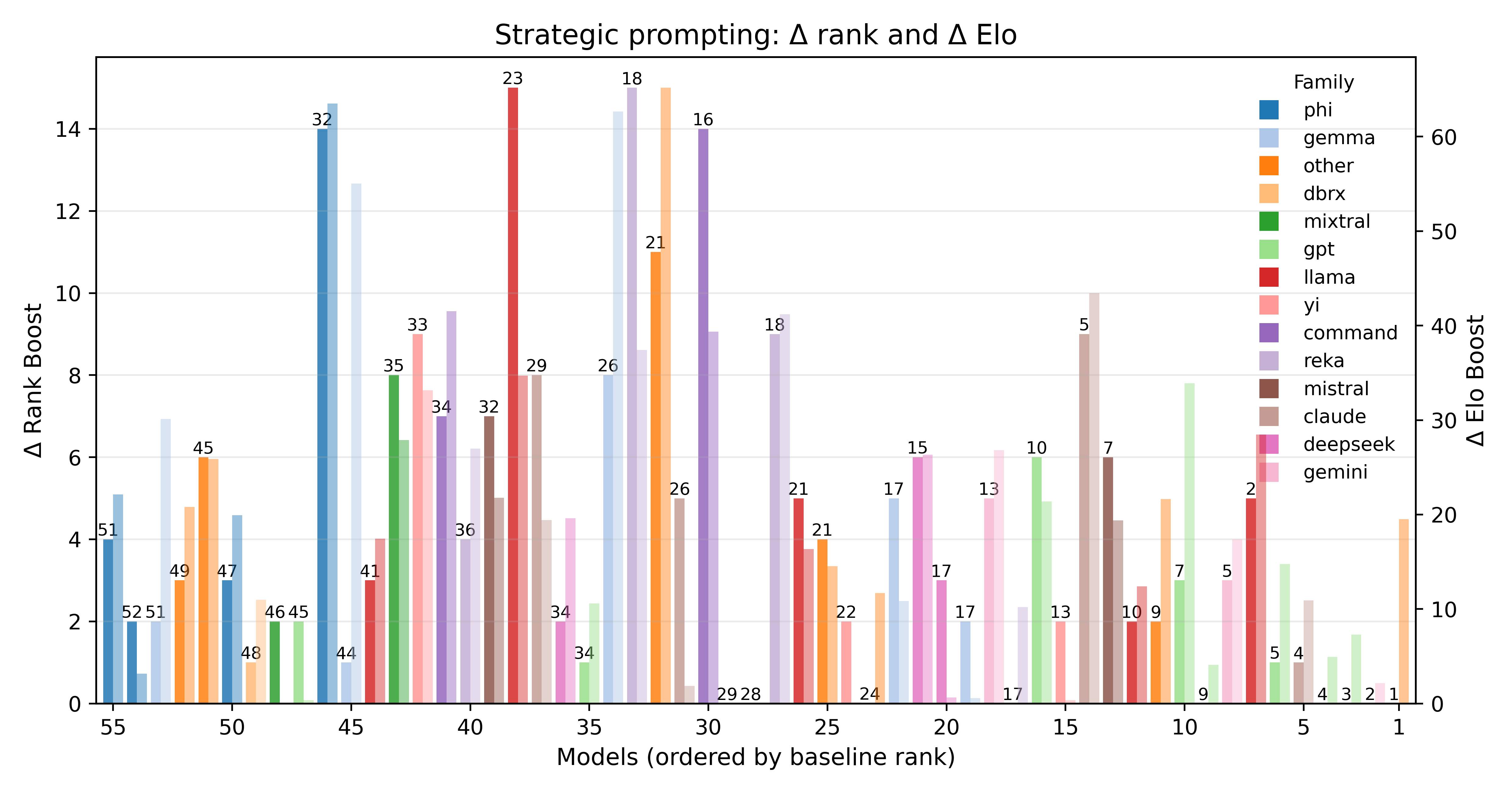}} 
    \caption{
      Change in model rank and Elo achieved by adding strategic prompts, capped at 10K prompts ($\sim10\%$ of total battles). Models are boosted up to 15 places  in the ranking.
    }
    \label{fig:prompt_rank_elo_change}
  \end{center}
\end{figure*}

Strategic prompting boosts model ranks almost as effectively as strategic voting: models move up to 15 places upward in the rankings, including moving several models up to the top 5 models (Figure \ref{fig:prompt_rank_elo_change}). For example, \texttt{reka-core-20240501} moves from 33rd place to 18th place by prompting in Hungarian, \texttt{llama-3.1-8b-instruct} moves from 38th place to 23rd place by prompting in Croatian; and \texttt{phi-3-medium-4k-instruct} moves from 46th place to 32nd place by prompting in Scots. This method is particularly effective with small numbers of prompts, moving model ranks with as few as 500 strategic prompts added (Figure \ref{fig:prompt_rank_change_box}, Appendix \ref{app:model-family}).

In principle, leaderboard developers could  spot and filter out unusual clusters of prompts in a language. However, the use of ``language'' as the axis for strategic prompting is arbitrary. This strategy works for any definable subcategories, meaning it is easy to pick subcategories that are harder to spot. As a demonstration (Appendix \ref{sec:prompting-other-subcats}), we repeated the prompt experiments with prompts split into subcategories based on whether they involve complexity, creativity, domain knowledge, and/or math.\footnote{These are preexisting LMArena labels, defined by having a model classify whether a prompt has each of these traits.} Under this version of strategic prompting, we see rank boosts of up to 14 places, comparable to those from strategic prompting using language for prompt subcategories.


\subsection{Combined Nomination and Prompt Attacks}
\begin{figure*}
  \begin{center}
    \centerline{\includegraphics[width=0.67\textwidth]{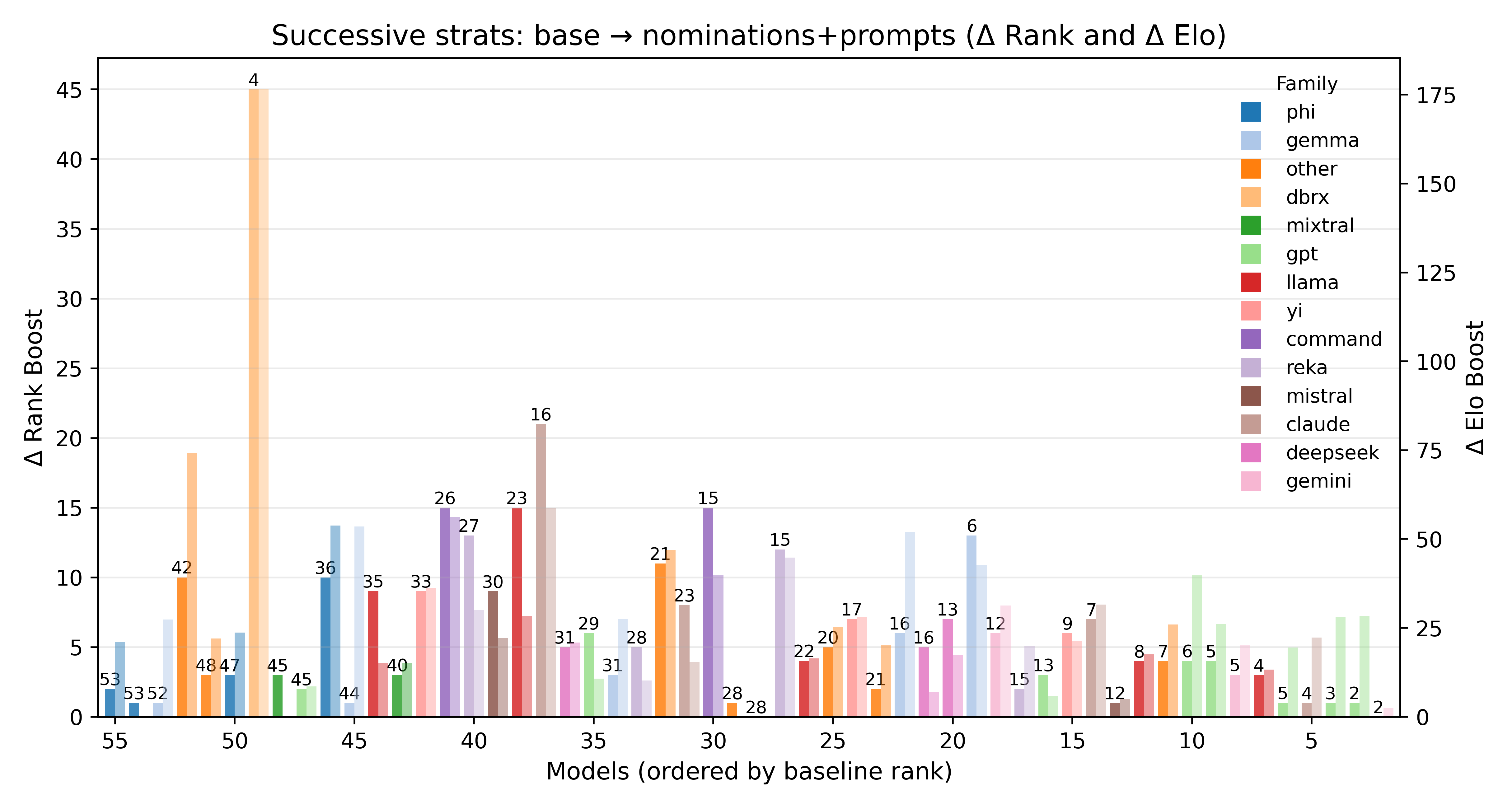}}
    \caption{
      Change in rank and Elo when performing strategic nominations followed by strategic prompting. Models are boosted up to 45 places in the rankings, with 9 models moving over 10 places upwards.
    }
    \label{fig:successive_noms_prompts}
  \end{center}
  \vspace{-4mm}
\end{figure*}

We examine the effects of combining strategic nomination and prompting (Figure \ref{fig:successive_noms_prompts}). To combine the attacks, for each model, we follow the strategic nomination strategy to add copies of the most favorable model; then, we run the strategic prompting algorithm on the dataset with those model copies added. We find that \textbf{models are boosted up to 45 places in the ranking}, with 9 models moving over 10 places upwards.


\subsection{Factors Contributing to Ease of Manipulation}
\label{sec:regressions}
The results across strategic voting, nomination, and prompting indicate that \textbf{some models are much easier to move than others}. For example, whereas a few models only move 1-2 ranks under all strategies, \texttt{dbrx-instruct-preview} moves from 49th to 39th place with strategic nominations; \texttt{command-r} moves from 41st to 21st place under strategic voting; and \texttt{reka-core-20240501} moves from 33rd to 18th place under strategic prompting.

We run regressions to understand what factors make it easier to manipulate a model's rank (Appendix \ref{app:regressions}). The factors we consider are $A$’s variance in win rate against different models; $A$’s starting rank and Elo; the number of models in $A$’s model family; $A$'s win rate, gap in Elo, and total battles against the model one rank higher; the number of battles in $A$'s best language for strategic prompting, and $A$'s number of total battles and percent of tied battles. To capture if the model could easily swap places with a model close above it, we define the ``Elo density'' as the number of models with Elo 0-50 points higher than $A$. For strategic voting ($R^2$=0.660), the top feature is Elo density, followed by win rate variance. For strategic prompting ($R^2$=0.518), the top features are the number of battles in the best prompting language and the gap in Elo score vs. the model one rank higher. For strategic nomination ($R^2$=0.557), the top features are Elo density and gap in Elo score vs. the model one rank higher. These features suggest that models placed below a cluster of models with slightly higher Elo scores are easier to shift in the rankings, since smaller changes in Elo result in larger changes in model rank.


\section{Attack Feasibility}\label{sec:feasibility}
\begin{table}[h]
\centering
\caption{Comparison of attack strategies and their properties.}
\renewcommand{\arraystretch}{1.3}
\fontsize{8.5}{10}\selectfont
\begin{tabular}{p{2cm}p{2cm}p{2.4cm}p{1.8cm}p{1.8cm}p{1.7cm}} %
\hline
\textbf{} & \textbf{Voting (greedy)} \textit{also see \citep{min2025improving}} & \textbf{Voting (heuristic)} & \textbf{Prompting} & \textbf{Nomination} & \textbf{Prompting + nomination} \\
\hline
\textbf{Max rank boost} & 20 places $\uparrow$ & 15 places $\uparrow$ & 15 places $\uparrow$ & 15 places $\uparrow$ & 45 places $\uparrow$ \\
\hline
\textbf{Must identify models?} & All models & Only target model $A$ & -- & -- & -- \\
\hline
\textbf{Cost of Elo calculations} ($m$ models, $n$ votes; \quad $n \gg m$) & $O(n)$ & $0$ & $\leq O(m)$; depends on desired \# of prompt subsets & $O(m)$ & $O(m)$ \\
\hline
\textbf{Possible \quad \quad \quad defenses} & Block bots \& prompting w/o voting & Block bots \& prompting w/o voting & Stratify/cluster prompts & Harmonic Borda count & Harmonic Borda count \\
& & &  & &  Stratify/cluster prompts \\
\hline
\end{tabular}
\vspace{-0.2cm}
\label{tab:attack-comparison}
\end{table}

\textbf{All attacks we describe are highly feasible}, as shown in Table \ref{tab:attack-comparison}. First, identifying the best way to boost a target model $A$ is computationally cheap.
Strategic nomination for a target model requires iterating once over a set of candidate models (55 in our dataset; 300 on the live version of LMArena). Strategic prompting requires iterating over a set of candidate subsets ($\sim$40 languages in our dataset; easy to restrict to the top $n$ languages on live data). Optimally computing which model to nominate or language to prompt in is a one-time computation per model; computing it for all 55 models in this dataset takes less than 5 minutes on one CPU. Greedy strategic voting on a random stream is more expensive (for each vote, the change in Elo if one model or the other were selected must be calculated), but using the heuristic strategy is much cheaper and the cost per vote does not depend on the number of models or battles in the dataset. In addition, as we have shown, identifying the models in an anonymized battle is relatively straightforward via multiple methods.

\textbf{How do the manipulations scale as dataset size increases?} We expect our manipulation effects to be feasible for LMArena's leaderboards that are similar in number of votes ($\sim$100K) to our dataset (e.g., code and search). However, LMArena's text leaderboard has $>$5 million votes; we expect that a higher absolute number of manipulated votes/prompts/nominations would be necessary to change the text rankings (limitations discussed further in Section \ref{app:limitations}). Our scaling experiments in Appendix \ref{app:scaling} show that the number of manipulations needed to boost a model \textbf{scales roughly linearly} with dataset size.
Though larger arenas are more secure, there is no exponential advantage to a larger leaderboard.


\section{Leaderboard Defenses}\label{sec:defenses}

While leaderboards currently implement some mitigations against strategic voting, such as bot detection \citep{huang2025exploring}, we do not know of any protections against strategic nomination (in which the attack is carried out by those submitting unrelated models) or strategic prompting (in which users' votes indeed align with their "true" preferences, but the prompts are skewed to favor the target model). We also find that properties of how the leaderboards are currently designed (e.g., ease of prompting to deanonymize models; ability to abstain from voting) facilitate the attacks we describe. Bearing in mind the potential impacts of studying leaderboard attacks (Section \ref{app:impacts}), we discuss operational and algorithmic defenses to protect against each attack we demonstrate:

\subsection{Strategic nomination defenses.}\label{sec:strat-nom-defense} Unlike strategic voting, strategic nomination involves manipulation by those submitting models to the leaderboard rather than those evaluating them. A simple defense against this method is to prevent the submission of many similar models at once. A more robust defense to both strategic voting and strategic nomination is to change the aggregation function of the Bradley-Terry model to put less weight on a voter $V$'s preferences among models that $V$ ranks poorly overall.

\paragraph{Harmonic Bradley-Terry objective.} We propose a defense that draws on a solution to the parallel problem in the social choice literature: whereas the Borda count (equivalent to the Bradley-Terry objective) gives $n-r$ points to the model at rank $r$ in a list of length $n$ (+1 point for each step it goes up in a voter's ranking), we can instead use the Dowdall or \textit{harmonic} Borda count, which gives $\frac{1}{r}$ points to the model at rank $r$. In real election settings, this change hampers strategic voting and makes strategic nomination require enough new nominations as to be broadly infeasible.\footnote{The government of Kiribati uses classic Borda count for some elections; Nauru uses the harmonic Dowdall variant. \citet{reilly2002} finds that strategic voting and nomination are both documented (and effective) in Kiribati, but he and \citet{Fraenkel03042014} find that while the same strategies are used in Nauru, they seem far less effective than under classic Borda count.} We propose a BT-compatible implementation of harmonic Borda  and find it reduces strategic nomination's effect on boosting Elo scores.

We rank the models based on their head-to-head win rate. Let $r(m)$ be the rank of model $m$ in this ranking. We assign the weights $w_{m} = \frac{1}{r(m)}$, which adds more weight to wins by models that perform well. We interpolate between the standard BT objective (all weights are 1) and the harmonic one using $w_{m, \beta} = (1 - \beta) + \beta w_{m}$. At $\beta=0$, every battle gets weight 1 (recovering standard BT); at $\beta=1$, battles are weighted only by the harmonic weight. This permits smooth tuning between the standard and harmonic BT objectives. We then reweight the matrix of pairwise wins by $w_{m, \beta}$ and aggregate using the standard BT objective (details in Appendix \ref{app:harmonic-borda}).

The harmonically-reweighted BT \textbf{reduces extreme boosts from strategic nomination}: whereas standard Borda count permits a maximum change in rank of 15 places, the harmonic variant reduces this to just 8 places (Figure \ref{fig:harm-borda}). However, the mean possible boost increases slightly (2.74 -> 2.81 places), indicating that small boosts to model rank across the board remain possible. Studying the effects of the harmonic Borda count on other electoral manipulation strategies in this setting, as well as its tradeoffs for leaderboards overall, remain important areas for future work.

\subsection{Strategic voting and strategic prompting defenses.} The ease of identifying models facilitates strategic voting. Since models are still readily identifiable through likelihood-based attribution, one defense is to check for voters who exhibit suspicious behavior, e.g. votes that always make one model's ranking improve on the leaderboard, since a true stream of votes on unrelated models should have mixed effects. This might require voters to be logged in and verified, to prevent rapid, anonymous manipulated votes from a single voter. Strategic voting would then be hampered because it would require many voters to coordinate. Finally, strategic voting is easiest when attackers need only vote on comparisons that involve the target model, which makes the strategic votes have a more concentrated effect. Thus, leaderboards should prevent users from viewing many comparisons without voting on them.
Strategic voting still remains difficult to detect if spread across a user group or if a user only manipulates some of their votes.

Strategic prompting allows individual votes to correctly align with broader preferences on that comparison, but shifts the prompt distribution so more prompts favorable to a particular model come up. When comparisons are clustered, model results become skewed. Approaches to address this could involve stratifying or clustering prompts and then weighting these groupings evenly so that overall performance is not unduly influenced by any single grouping (e.g., prompt language). As a theoretical contribution in Appendix \ref{sec:less-lm-arena-elo}, we propose a modified version of Bradley-Terry that replaces individual scores with kernel-smoothed scores to make it more difficult to use clusters of comparisons to manipulate model rankings.

\section{Conclusion}\label{sec:conclusion}
As model evaluations become targets for highly incentivized actors, such as model providers who benefit from improving their models' perceived performance, the likelihood and potential impact of ranking manipulations increases. We drew upon known weaknesses of the Borda count in social choice, which translates to the Bradley-Terry model used in most LLM leaderboards, and tested the equivalents of strategic voting, strategic nomination, and strategic prompting on LMArena data. We found that preference-based model rankings are vulnerable to all three attacks. Concerningly, these effects can move models substantially in the rankings and are relatively simple, cheap, and fast to compute and implement. We discussed defenses for each manipulation technique, including interface changes and variants of current aggregation techniques. Our findings highlight the need to secure widely used leaderboards against such attacks to ensure trust in preference-based model rankings.

\section*{Limitations}
\label{app:limitations}
The dataset we work with is smaller than the full LMArena rankings (though comparable to several of the domain-specific leaderboards). We expect that larger numbers of absolute manipulations would be necessary to move models on larger leaderboards. In addition, strategic voting in practice may be hampered if model identification is difficult; our main results assume the models can be identified. We also note that, for ethical reasons, we have not tried to move models on the live rankings themselves. Red-teaming for the effects of these sorts of manipulations on live rankings, and for effects on larger leaderboards, remain important areas for future work.

\para{Impact Statement}\label{app:impacts}
Research on potential methods for leaderboard ``hacking'' raises the concern that bad actors could use these methods in order to manipulate their model results. To mitigate this, we discuss defenses to each of these methods. Our goal is that, in making these simple methods widely known (rather than waiting for bad-faith actors to use them privately), this work raises awareness of the need to defend against manipulated evaluations.

\section*{Acknowledgments}
Many thanks to the members of the MilaNLP and Berkeley NLP groups for many helpful conversations, and to Chris Hays and Anastasios Angelopoulos for their generous and thoughtful insights. This work was partly supported by an NSF GRFP grant and the SNSF grant 235328.





\bibliographystyle{unsrtnat}
\bibliography{leaderboard_hacking.bib}

@inproceedings{siththaranjandistributional,
  title={Distributional Preference Learning: Understanding and Accounting for Hidden Context in RLHF},
  author={Siththaranjan, Anand and Laidlaw, Cassidy and Hadfield-Menell, Dylan},
  booktitle={The Twelfth International Conference on Learning Representations},
  year="2024"
}

@article{Fraenkel03042014,
author = {Jon Fraenkel and Bernard Grofman},
title = {The Borda Count and its real-world alternatives: Comparing scoring rules in Nauru and Slovenia},
journal = {Australian Journal of Political Science},
volume = {49},
number = {2},
pages = {186--205},
year = {2014},
publisher = {Routledge},
doi = {10.1080/10361146.2014.900530},
}

@inproceedings{hays2026strategiccandidacygenerativeai,
title={Strategic Candidacy in Generative {AI} Arenas},
author={Chris Hays and Rachel Li and Bailey Flanigan and Manish Raghavan},
booktitle={Forty-third International Conference on Machine Learning},
year={2026},
url={https://openreview.net/forum?id=K1mAvyHhKU}
}

@inproceedings{liu2025reevaluatingopenendedevaluationlarge,
title={Re-evaluating Open-ended Evaluation of Large Language Models},
author={Siqi Liu and Ian Gemp and Luke Marris and Georgios Piliouras and Nicolas Heess and Marc Lanctot},
booktitle={The Thirteenth International Conference on Learning Representations},
year={2025},
url={https://openreview.net/forum?id=kbOAIXKWgx}
}

@article{reilly2002,
author = {Benjamin Reilly},
title ={Social Choice in the South Seas: Electoral Innovation and the Borda Count in the Pacific Island Countries},

journal = {International Political Science Review},
volume = {23},
number = {4},
pages = {355-372},
year = {2002},
doi = {10.1177/0192512102023004002},

URL = { 
        https://doi.org/10.1177/0192512102023004002
},
eprint = {  
        https://doi.org/10.1177/0192512102023004002
}
}

@inproceedings{min2025improving,
title={Improving Your Model Ranking on Chatbot Arena by Vote Rigging},
author={Rui Min and Tianyu Pang and Chao Du and Qian Liu and Minhao Cheng and Min Lin},
booktitle={Forty-second International Conference on Machine Learning},
year={2025},
url={https://openreview.net/forum?id=5cDc71jLc1}
}

@inproceedings{zhao2025challenges,
  title={Challenges in trustworthy human evaluation of chatbots},
  author={Zhao, Wenting and Rush, Alexander M and Goyal, Tanya},
  booktitle={Findings of the Association for Computational Linguistics: NAACL 2025},
  pages={3359--3365},
  year={2025}
}

@article{huang2025exploring,
  title={Exploring and mitigating adversarial manipulation of voting-based leaderboards},
  author={Huang, Yangsibo and Nasr, Milad and Angelopoulos, Anastasios and Carlini, Nicholas and Chiang, Wei-Lin and Choquette-Choo, Christopher A and Ippolito, Daphne and Jagielski, Matthew and Lee, Katherine and Liu, Ken Ziyu and others},
  journal={arXiv preprint arXiv:2501.07493},
  year={2025}
}

@misc{bai2022traininghelpfulharmlessassistant,
      title={Training a Helpful and Harmless Assistant with Reinforcement Learning from Human Feedback}, 
      author={Yuntao Bai and Andy Jones and Kamal Ndousse and Amanda Askell and Anna Chen and Nova DasSarma and Dawn Drain and Stanislav Fort and Deep Ganguli and Tom Henighan and Nicholas Joseph and Saurav Kadavath and Jackson Kernion and Tom Conerly and Sheer El-Showk and Nelson Elhage and Zac Hatfield-Dodds and Danny Hernandez and Tristan Hume and Scott Johnston and Shauna Kravec and Liane Lovitt and Neel Nanda and Catherine Olsson and Dario Amodei and Tom Brown and Jack Clark and Sam McCandlish and Chris Olah and Ben Mann and Jared Kaplan},
      year={2022},
      eprint={2204.05862},
      archivePrefix={arXiv},
      primaryClass={cs.CL},
      url={https://arxiv.org/abs/2204.05862}, 
}

@inproceedings{boubdir2024,
 author = {Boubdir, Meriem and Kim, Edward and Ermis, Beyza and Hooker, Sara and Fadaee, Marzieh},
 booktitle = {Advances in Neural Information Processing Systems},
 doi = {10.52202/079017-3367},
 pages = {106135--106161},
 publisher = {Curran Associates, Inc.},
 title = {Elo Uncovered: Robustness and Best Practices in Language Model Evaluation},
 url = {https://proceedings.neurips.cc/paper_files/paper/2024/file/bfba8efb806a970455b83b852c9cf846-Paper-Conference.pdf},
 volume = {37},
 year = {2024}
}

@inproceedings{singh2025leaderboardillusion,
 author = {Singh, Shivalika and Nan, Yiyang and Wang, Alex and Dsouza, Daniel and Kapoor, Sayash and \"{U}st\"{u}n, Ahmet and Koyejo, Sanmi and Deng, Yuntian and Longpre, Shayne and Smith, Noah and Ermis, Beyza and Fadaee, Marzieh and Hooker, Sara},
 booktitle = {Advances in Neural Information Processing Systems},
 doi = {10.52202/085713-2620},
 pages = {},
 publisher = {Curran Associates, Inc.},
 title = {The Leaderboard Illusion},
 url = {https://proceedings.neurips.cc/paper_files/paper/2025/file/70a93f260a51123b3c0e33ecd1b4de97-Paper-Datasets_and_Benchmarks_Track.pdf},
 volume = {38, Main Conference},
 year = {2025}
}

@inproceedings{chiang2024,
author = {Chiang, Wei-Lin and Zheng, Lianmin and Sheng, Ying and Angelopoulos, Anastasios N. and Li, Tianle and Li, Dacheng and Zhu, Banghua and Zhang, Hao and Jordan, Michael I. and Gonzalez, Joseph E. and Stoica, Ion},
title = {Chatbot arena: an open platform for evaluating LLMs by human preference},
year = {2024},
publisher = {JMLR.org},
booktitle = {Proceedings of the 41st International Conference on Machine Learning},
articleno = {331},
numpages = {30},
location = {Vienna, Austria},
series = {ICML'24}
}

@book{brandt2016handbook,
  title={Handbook of computational social choice},
  author={Brandt, Felix and Conitzer, Vincent and Endriss, Ulle and Lang, J{\'e}r{\^o}me and Procaccia, Ariel D},
  year={2016},
  publisher={Cambridge University Press}
}

@article{sadasivan2025can,
title={Can {AI}-Generated Text be Reliably Detected? Stress Testing {AI} Text Detectors Under Various Attacks},
author={Vinu Sankar Sadasivan and Aounon Kumar and Sriram Balasubramanian and Wenxiao Wang and Soheil Feizi},
journal={Transactions on Machine Learning Research},
issn={2835-8856},
year={2025},
url={https://openreview.net/forum?id=OOgsAZdFOt},
}

@inproceedings{verma-etal-2024-ghostbuster,
    title = "Ghostbuster: Detecting Text Ghostwritten by Large Language Models",
    author = "Verma, Vivek  and
      Fleisig, Eve  and
      Tomlin, Nicholas  and
      Klein, Dan",
    booktitle = "Proceedings of the 2024 Conference of the North American Chapter of the Association for Computational Linguistics: Human Language Technologies (Volume 1: Long Papers)",
    month = jun,
    year = "2024",
    address = "Mexico City, Mexico",
    publisher = "Association for Computational Linguistics",
    url = "https://aclanthology.org/2024.naacl-long.95/",
    doi = "10.18653/v1/2024.naacl-long.95",
    pages = "1702--1717"
}

@inproceedings{uchendu-etal-2020-authorship,
    title = "Authorship Attribution for Neural Text Generation",
    author = "Uchendu, Adaku  and
      Le, Thai  and
      Shu, Kai  and
      Lee, Dongwon",
    booktitle = "Proceedings of the 2020 Conference on Empirical Methods in Natural Language Processing (EMNLP)",
    month = nov,
    year = "2020",
    address = "Online",
    publisher = "Association for Computational Linguistics",
    url = "https://aclanthology.org/2020.emnlp-main.673/",
    doi = "10.18653/v1/2020.emnlp-main.673",
    pages = "8384--8395"
}

@InProceedings{pmlr-v235-zhang24u,
  title = 	 {Inherent Trade-Offs between Diversity and Stability in Multi-Task Benchmarks},
  author =       {Zhang, Guanhua and Hardt, Moritz},
  booktitle = 	 {Proceedings of the 41st International Conference on Machine Learning},
  pages = 	 {58984--59002},
  year = 	 {2024},
  volume = 	 {235},
  series = 	 {Proceedings of Machine Learning Research},
  month = 	 {21--27 Jul},
  publisher =    {PMLR},
  url = 	 {https://proceedings.mlr.press/v235/zhang24u.html}
}

@inproceedings{chen2026leaderboard,
title={Leaderboard Incentives: Model Rankings under Strategic Post-Training},
author={Yatong Chen and Guanhua Zhang and Moritz Hardt},
booktitle={Forty-third International Conference on Machine Learning},
year={2026},
url={https://openreview.net/forum?id=r6wfuAKmVb}
}

\newpage
\appendix
\onecolumn
\section{Harmonic Bradley-Terry Experiments}\label{app:harmonic-borda}
\begin{figure*}
  \begin{center}
    \centerline{\includegraphics[width=0.55\textwidth]{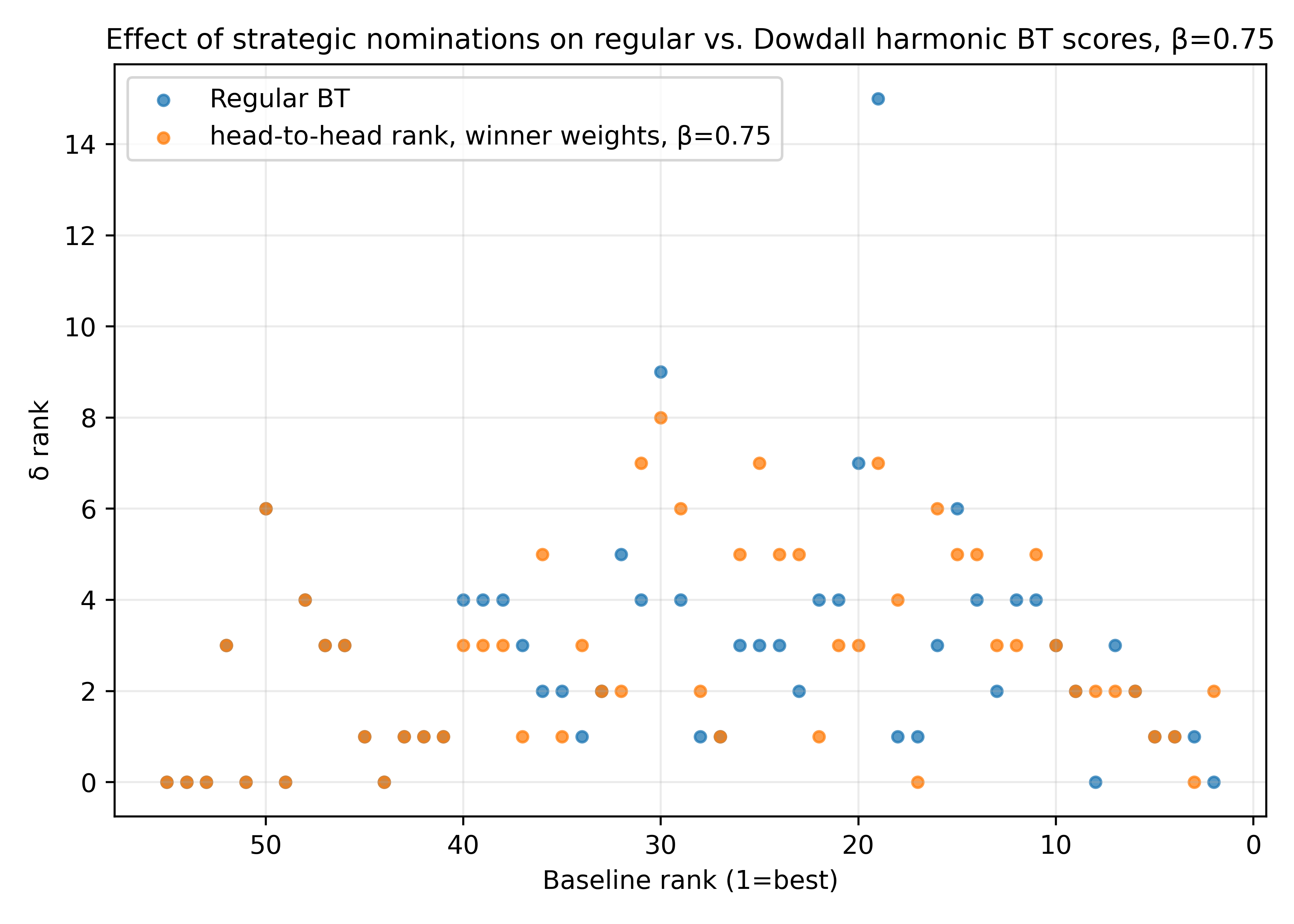}}
    \caption{
      Boost in rank achievable under strategic nomination when using the standard Borda count (blue) vs. the harmonically weighted Borda count (orange). The harmonic version reduces the maximum boost from 15 places to 8 places.
    }
    \label{fig:harm-borda}
  \end{center}
  \vspace{-4mm}
\end{figure*}

We propose an implementation of a harmonically-weighted Borda count that maintains compatibility with the Bradley-Terry objective. The Borda count gives $n-r$ points to the model at rank $r$ in a list of length $n$ (+1 point for each model it beats in a voter's ranking). The Dowdall or \textit{harmonic} Borda count instead gives $\frac{1}{r}$ points to the model at each voter's rank $r$.

For the harmonic Borda count, we first compute a ranking of models based on the number of models that they beat head-to-head, then assign a weight $1/r$ to each model at rank $r$ in this ranking.
We interpolate between the standard BT objective (all weights are 1) and the harmonic one using $w_{m, \beta} = (1 - \beta) + \beta w_{m}$. We experiment with $\beta = [0.25, 0.5, 0.75, 1]$. Since switching to harmonic Borda count also changes the original rankings, we measure the extent to which harmonic Borda count modifies the original ranking using the Kendall-Tau distance. We find that $\beta=0.75$ gives the best tradeoff of minimizing extreme boosts and distance from the original ranking (Table \ref{tab:harmonic-beta}).

\begin{table}[h!]
\centering
\begin{tabular}{cccc}
\hline
$\beta$ & Mean best boost & Max boost & Kendall $\tau$ with regular BT \\
\hline
0    & 2.74 & 15 & 1    \\
0.25 & 2.74 & 14 & 0.993 \\
0.50 & 2.63 & 10 & 0.974 \\
\textbf{0.75} & \textbf{2.81} & \textbf{8}  & \textbf{0.946} \\
1.00 & 4.39 & 12 & 0.852 \\
\hline
\end{tabular}
\caption{Effect of harmonic Borda count on the highest boosts that strategic nomination can achieve (mean highest boost per model, and highest boost across models), as well as Kendall $\tau$ distance from original ranking, at different $\beta$ weightings.}
\label{tab:harmonic-beta}
\end{table}

In a variant of this method, we can interpret the harmonic Borda to use the ranking for \textit{voter sets} instead of individual voters. This is a closer equivalence to the harmonic Borda count for Bradley-Terry, but introduces additional challenges: the definition of harmonic Borda hinges on the notion of a complete ranking of all candidates for each voter (so that models can be reweighted based on their rank in that voter's ballot). But typical preference learning datasets do not contain a complete ranking per voter. Instead, we interpret the harmonic Borda to use the ranking for \textit{voter sets} instead of individual voters.

In this version, we require voter sets to be large enough to contain preferences involving all models in the dataset. Each voter set, consisting of the union of voters' preferences as edges, must be a connected comparison graph. We partition the judges into clusters that maximize the number of clusters while keeping the comparison graphs connected. We require voter sets to be large enough to contain preferences involving all models in the dataset. That is: for a set of models $m_1 \dots m_n$ and preferences $(m_i, m_j)$, each voter defines a comparison graph with models as the vertices and the voter's preferences as the edges. Each voter set, consisting of the union of voters' edges, must be a connected comparison graph with vertices $m_1 \dots m_n$. We partition the judges into clusters that maximize the number of clusters while keeping the comparison graphs connected. If the number of clusters is set to 1, this variant reduces to the harmonic Borda count version we discussed in Section \ref{sec:strat-nom-defense}.

In practice, we found that larger numbers of clusters--closer to the notion of reranking based on a harmonic reweighting of individual voters' rankings--tend to result in more unstable rankings because each cluster contains fewer datapoints. However, this strategy may be more feasible on larger datasets, a potential area for future work.

\section{Density-smoothed Bradley--Terry / Elo for LM-arena-style model rankings}
\label{sec:less-lm-arena-elo}

We describe a simple modification of standard Bradley--Terry (BT) / Elo fitting for pairwise preference data.
The modification replaces point scores with kernel-smoothed (KDE-like) ``regional mass'' scores.
The intent is to make repeated or highly clustered comparisons produce diminishing returns by coupling similar prompt--response instances through a similarity kernel.

We focus on model-level Elo parameters (one scalar per model), using LM-arena / LMR-style data: each comparison is between two model responses to a prompt.

\subsection{Standard Bradley--Terry / Elo (model-level)}
\label{subsec:standard-bt-elo}

\paragraph{Data and notation.}
A dataset $D$ consists of $N$ pairwise comparisons.
Each comparison $i$ has:
\begin{itemize}
  \item prompt $p_i$,
  \item model $m_i^A$ producing response $r_i^A$,
  \item model $m_i^B$ producing response $r_i^B$,
  \item label $y_i\in\{0,1\}$, where $y_i=1$ means $A$ is preferred to $B$.
\end{itemize}
We write the row as $(p_i, m_i^A, r_i^A;\; m_i^B, r_i^B;\; y_i)$.

In model-level Elo/BT, the learned parameter is one scalar per model:
\[
\theta_m \in \mathbb{R}, \qquad m\in\{1,\dots,M\}.
\]
Define the positive model strength
\[
w_m \;=\; \exp(\theta_m).
\]

\paragraph{Choice model.}
Standard Bradley--Terry assumes the probability that $A$ beats $B$ depends only on the two models:
\begin{align}
P_i^{BT}
&:= \Pr(m_i^A \succ m_i^B)
= \frac{w_{m_i^A}}{w_{m_i^A}+w_{m_i^B}}
\\ &= \sigma\!\big(\theta_{m_i^A}-\theta_{m_i^B}\big),
\end{align}
where $\sigma(t)=1/(1+e^{-t})$ is the logistic function.

\paragraph{Objective.}
The standard regularized negative log-likelihood is
\begin{align}
L_{BT}(\theta)
&= -\sum_{i=1}^N \Big[y_i\log P_i^{BT} + (1-y_i)\log\big(1-P_i^{BT}\big)\Big]
\\ &+ \frac{\lambda}{2}\sum_{m=1}^M \theta_m^2 .
\end{align}
The fitted $\theta$ values are then reported (after an affine rescaling) as Elo.

\paragraph{Why repetition has linear leverage.}
In standard BT/Elo, each match depends only on $(\theta_{m_i^A},\theta_{m_i^B})$.
Repeating the same comparison adds the same loss term again; influence is essentially linear in the amount of repetition (until regularization or finite-data effects dominate).

\subsection{Density-smoothed (kernel) Bradley--Terry / Elo}
\label{subsec:kernel-bt-elo}

\paragraph{Key change: ``items'' are prompt--response instances.}
Define the two \emph{items} compared in row $i$ as the prompt--response pairs
\[
x_i^A = (p_i, r_i^A),
\qquad
x_i^B = (p_i, r_i^B).
\]
Let $\mathcal{I}$ be the set of all unique items $x=(p,r)$ that appear in the dataset, and let $T=|\mathcal{I}|$.
Each item $x_t\in\mathcal{I}$ is produced by some model $m(t)$.

\paragraph{Kernel on embeddings.}
Define a similarity kernel between items
\[
K(x,z)\ge 0,
\]
where $x$ and $z$ are prompt--response pairs.
Intuitively, $K$ is large when two prompt--response pairs are similar (e.g., same/similar prompt and similar response) and small otherwise.
Practically, $K$ would be computed from embeddings of prompts/responses or a joint (prompt+response) embedding.

\paragraph{Kernel-smoothed (regional mass) score.}
Define the kernel-smoothed score (regional mass) for item $x$ as
\begin{align}
\tilde{s}(x;\theta)
&= \sum_{z\in \mathcal{I}_{multiset}} \exp(\theta_{m(z)})K(x,z).
\end{align}
Here $\mathcal{I}_{\mathrm{multiset}}$ is the multiset of items as they appear in the dataset: if an item $z$ appears multiple times across comparisons (on either side), it appears multiple times in the sum.

\paragraph{BT probability, but using smoothed scores.}
For comparison $i$, we have
\begin{align}
P_i^{KBT}
&:= \Pr(x_i^A \succ x_i^B)
= \frac{\tilde{s}(x_i^A;\theta)}{\tilde{s}(x_i^A;\theta)+\tilde{s}(x_i^B;\theta)}.
\end{align}
This has the same algebraic form as BT, but the scores are kernel-smoothed sums over (prompt, response) instances in the dataset.

\paragraph{Objective.}
Use the same Bernoulli log-loss form:
\begin{align}
L_{KBT}(\theta)
&= -\sum_{i=1}^N \Big[y_i\log P_i^{KBT} \\ &+ (1-y_i)\log\big(1-P_i^{KBT}\big)\Big]\\
&+ \frac{\lambda}{2}\sum_{m=1}^M \theta_m^2 .
\end{align}

\paragraph{Qualitative difference from standard BT.}
In standard BT, each match uses $(w_{m_i^A},w_{m_i^B})$ directly.
In kernel-smoothed BT, each match uses KDE-like regional masses that sum model scores over \emph{similar} prompt--response items across the dataset.
Therefore, adding (or repeating) a comparison changes the loss by adding another term, but also changes the \emph{scores used by other terms} via the kernel sums.
Evidence tends to accumulate at the level of regions of similarity, rather than enabling arbitrarily large separation among near-duplicate items within the same region.

\paragraph{Implementation simplification for per-model $\theta$.}
Because $\mathrm{score}(z)$ depends only on the model identity of $z$, each smoothed score is a fixed linear combination of $\exp(\theta_m)$:
\[
w_m = \exp(\theta_m).
\]
Define weights
\begin{align}
A_{tm}
&:= \sum_{z\in\mathcal{I}_{multiset}:\,m(z)=m} K(x_t,z).
\end{align}
Then for item $t$,
\begin{align}
\tilde{s}(x_t;\theta)
&= \sum_{m=1}^M A_{tm}\,w_m.
\end{align}
We note that, as a limitation, computing $A$ may be expensive. However, once $A$ is computed, evaluating a single match probability only requires two dot products with $w$:
\[
P_i^{KBT} = \frac{A_{t_i^A}\cdot w}{\left(A_{t_i^A} + A_{t_i^B}\right) \cdot w}.
\]
Thus, after precomputing $A$, SGD-style optimization looks very similar to standard BT/Elo: each (mini)batch  only requires the rows of $A$ corresponding to the items in that batch.

\subsection{Simple limiting behavior and intuition}
\label{subsec:limiting-behavior}

\paragraph{Duplicates of a single comparison (same prompt, same responses).}
Suppose we repeatedly add copies of one labeled comparison $(p, a \succ b)$.
In standard BT, this adds repeated identical terms and pushes $\theta_{m(a)}-\theta_{m(b)}$ upward roughly in proportion to repetition (until regularized).

In kernel-smoothed BT, repeating the comparison increases the regional mass around both $x=(p,a)$ and $x=(p,b)$ (and around nearby items) because both sides appear in the kernel sums.
This yields diminishing returns: additional duplicates also increase the denominator of related comparisons through the KDE. This counteracts the incentive to increase the model score, since increases will also push nearby comparisons closer together. 

\paragraph{Duplicates concentrated on a single model.}
An ``attack'' that makes one model appear and win many times is essentially linear in standard BT: more wins leads to more score movement.
In the kernel-smoothed objective, concentrated wins inflate local regional mass in the neighborhoods where those wins occur.
If the wins are highly clustered (same prompt type or response style), the marginal gain in moving $\theta$ diminishes.
Large global movement requires that the wins be spread across diverse regions of the prompt--response space.

\paragraph{Duplicates concentrated in a single language.}
If the kernel uses prompt similarity (or prompt+response embeddings), then a single language will typically form a coherent region in embedding space.
This means that a large number of comparisons in a single language will dilute the effect, because there will also be a large contribution from that model to the denominator.

\label{app:less-estimation}

\section{Results: Supporting Details}
\label{app:model-family}

\begin{figure}
  \vskip 0.2in
  \begin{center}
    \centerline{\includegraphics[width=0.6\columnwidth]{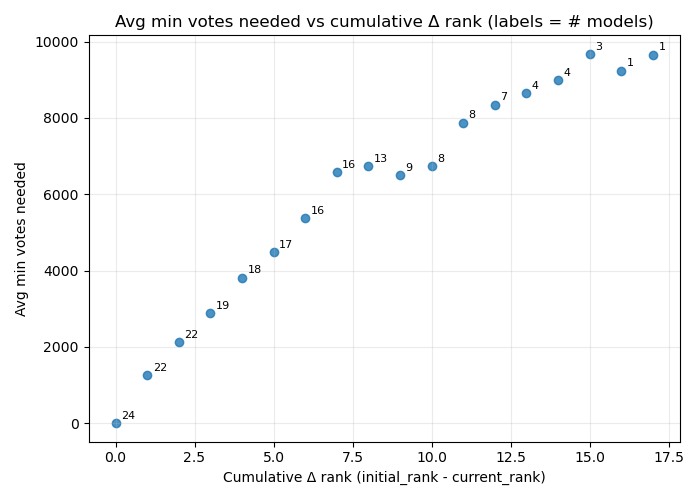}}
    \centerline{\includegraphics[width=0.6\columnwidth]{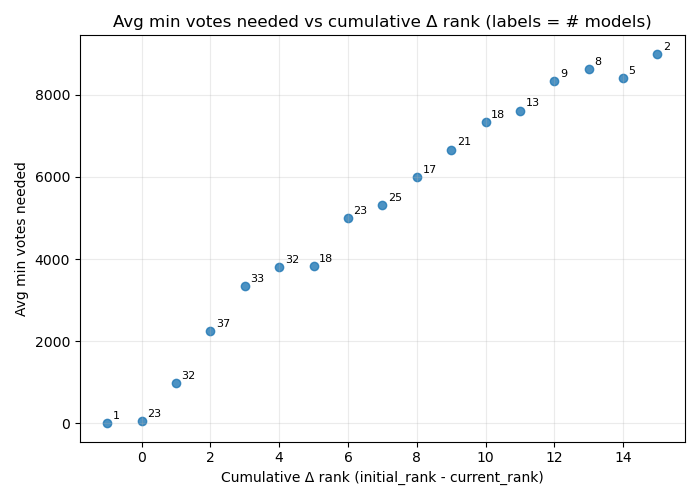}}
    \caption{
      Number of votes under the \textbf{greedy strategy (top)} and \textbf{heuristic strategy (bottom)}  needed to achieve greater improvements in leaderboard rank. Point labels indicate the number of models that can achieve that rank improvement with under 10K strategic votes. Small improvements to rank (2-3 places) are relatively easy to achieve, requiring under 2\% of total votes (2K of 100K total) to be manipulated and achievable by a relatively large subset of models. Larger improvements (15+ places) are achievable by a smaller subset of models with more manipulated votes. While the maximum rank boost achieved is higher for the greedy strategy, more models are able to achieve smaller boosts under the heuristic strategy (and typically with fewer votes).
    }
    \label{fig:votes_needed_for_mvt}
  \end{center}
\end{figure}

Figure \ref{fig:votes_needed_for_mvt} tracks the number of votes needed to move a model up a certain number of places in the rankings under greedy vs. heuristic voting strategies.

\begin{figure}
  \vskip 0.2in
  \begin{center}
    \centerline{\includegraphics[width=0.5\columnwidth]{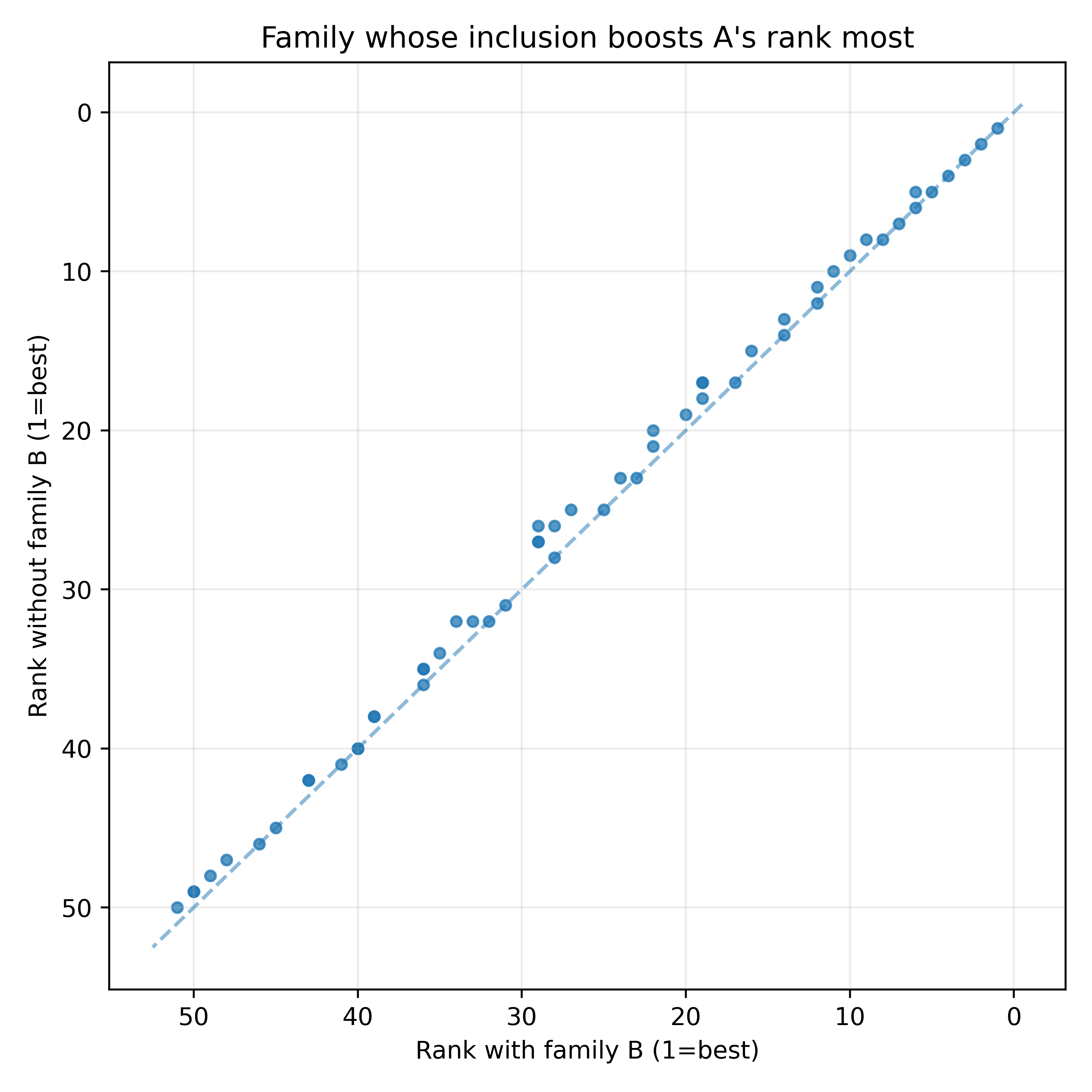}}
    \caption{
      Counterfactual of model ranks with and without the family $B$ of models that moves the target model furthest \textit{up} when included. We plot each model's rank when family $B$ is excluded from the Elo calculation, versus its rank (among models not in $B$) when family $B$ is included in the Elo dataset, i.e., the original dataset.
    }
    \label{fig:family-removal-noms-up}
  \end{center}
\end{figure}

\begin{figure}
  \vskip 0.2in
  \begin{center}
    \centerline{\includegraphics[width=0.5\columnwidth]{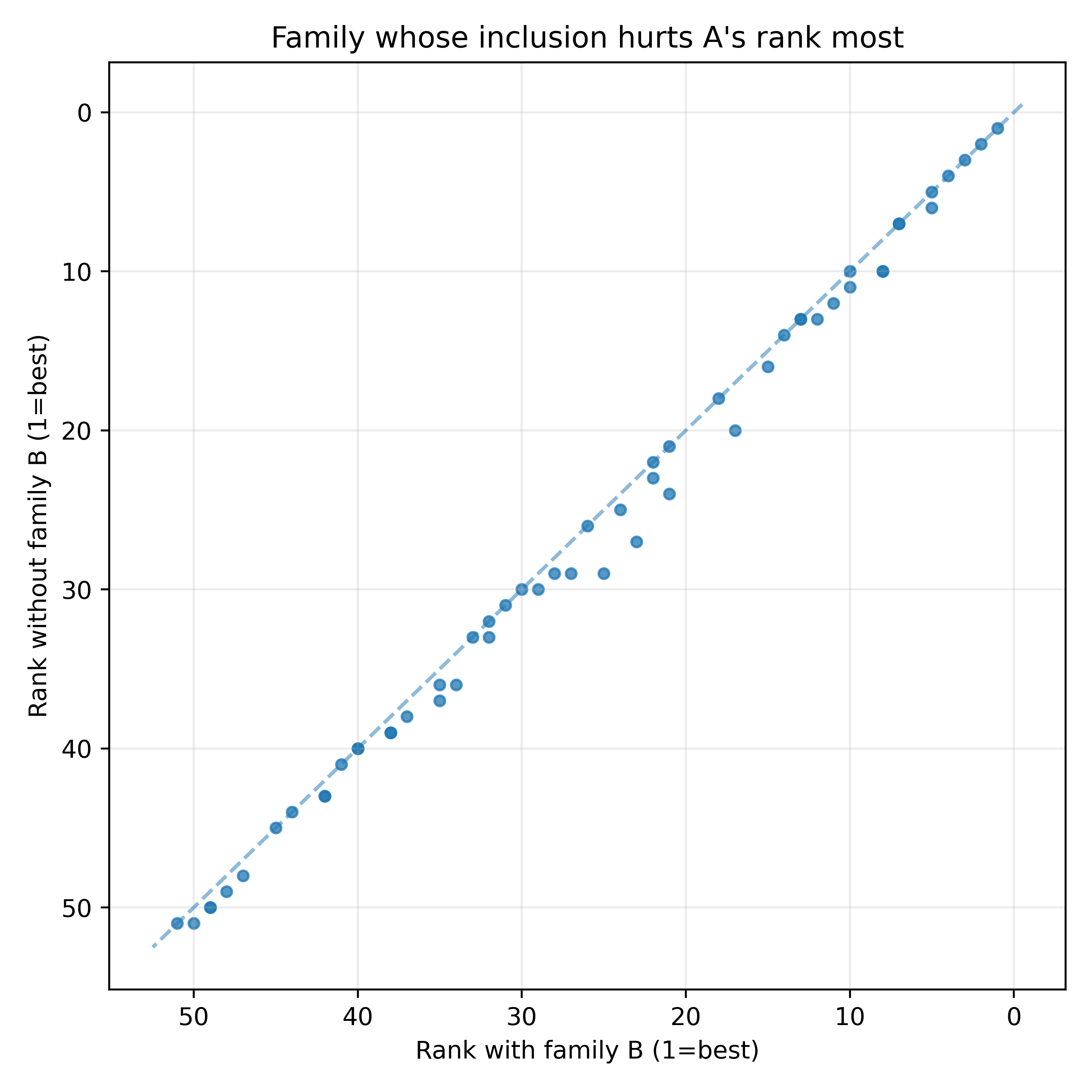}}
    \caption{
      Counterfactual of model ranks with and without the family $B$ of models that moves the target model furthest \textit{down} when included. We plot each model's rank when family $B$ is excluded from the Elo calculation, versus its rank (among models not in $B$) when family $B$ is included in the Elo dataset, i.e., the original dataset.
    }
    \label{fig:family-removal-noms-down}
  \end{center}
\end{figure}

Figures \ref{fig:family-removal-noms-up} and \ref{fig:family-removal-noms-down} indicate the boost achieved in a model's rank by adding the model family that favors it most. Inclusion of the most favorable family can move models 1-4 places upward, if desired, or 1-4 places downward.

\begin{figure}[!ht]
  \vskip 0.2in
  \begin{center}
    \centerline{\includegraphics[width=\columnwidth]{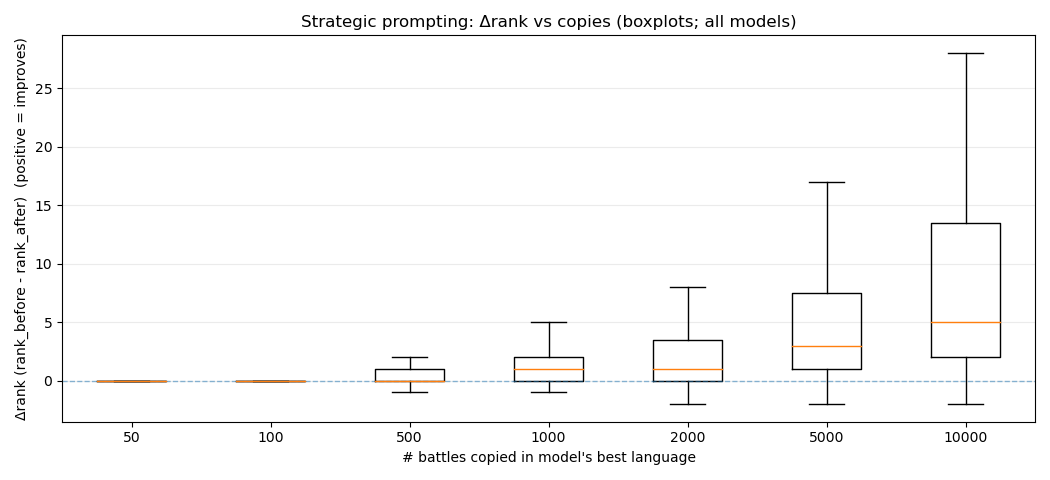}}
    \caption{
      Average rank change achieved as more strategic prompts are added.
    }
    \label{fig:prompt_rank_change_box}
  \end{center}
\end{figure}

Figure \ref{fig:prompt_rank_change_box} indicates the number of strategic prompts needed to achieve different changes in model rank.

\section{Live model identification}
\label{app:live_model_identification}

We evaluate live model identification on the \url{lmarena-ai/arena-human-preference-100k} dataset.
We consider seven open-weight models and three proprietary API-based models.
The evaluated models include:
\begin{itemize}[nosep]
    \item \texttt{gpt-4o-2024-08-06}
    \item \texttt{gpt-4o-mini-2024-07-18}
    \item \texttt{gpt-3.5-turbo-0125}
    \item \texttt{llama-3.1-70b-instruct}
    \item \texttt{llama-3-70b-instruct}
    \item \texttt{llama-3-8b-instruct}
    \item \texttt{llama-3.1-8b-instruct}
    \item \texttt{qwen2-72b-instruct}
    \item \texttt{gemma-2-2b-it}
    \item \texttt{gemma-1.1-7b-it}
\end{itemize}
For each model, we sample $N=200$ prompt-response pairs, of which 25\% have been generated by the target model and the remaining 75\% by other models.
We then formulate the identification as a ranking task, where we make a guess for $K$ outputs and we report Precision@$K$, where $K/N$ corresponds to the fraction of total outputs guessed.

\begin{figure}[thb]
  \vskip 0.2in
  \begin{center}
    \centerline{\includegraphics[width=0.6\columnwidth]{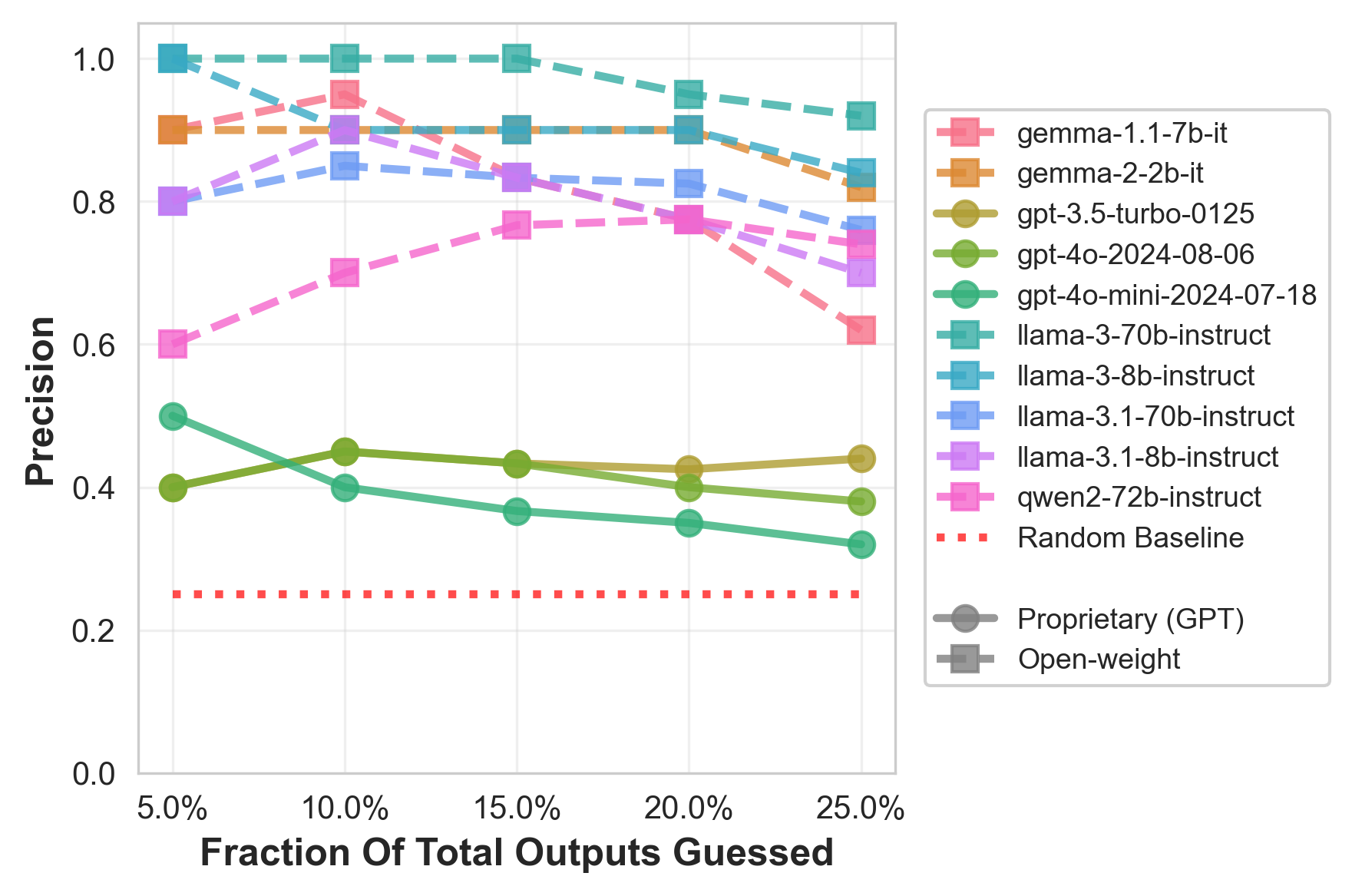}}
    \caption{
      Model detection precision at different fractions of total outputs guessed for open-weight models (square markers, dashed lines) and proprietary GPT models (circle markers, solid lines). Random baseline shown at 0.25 (base rate).
    }
    \label{live_model_identification}
  \end{center}
\end{figure}

\begin{figure}[thb]
  \vskip 0.2in
  \begin{center}
    \centerline{\includegraphics[width=0.6\columnwidth]{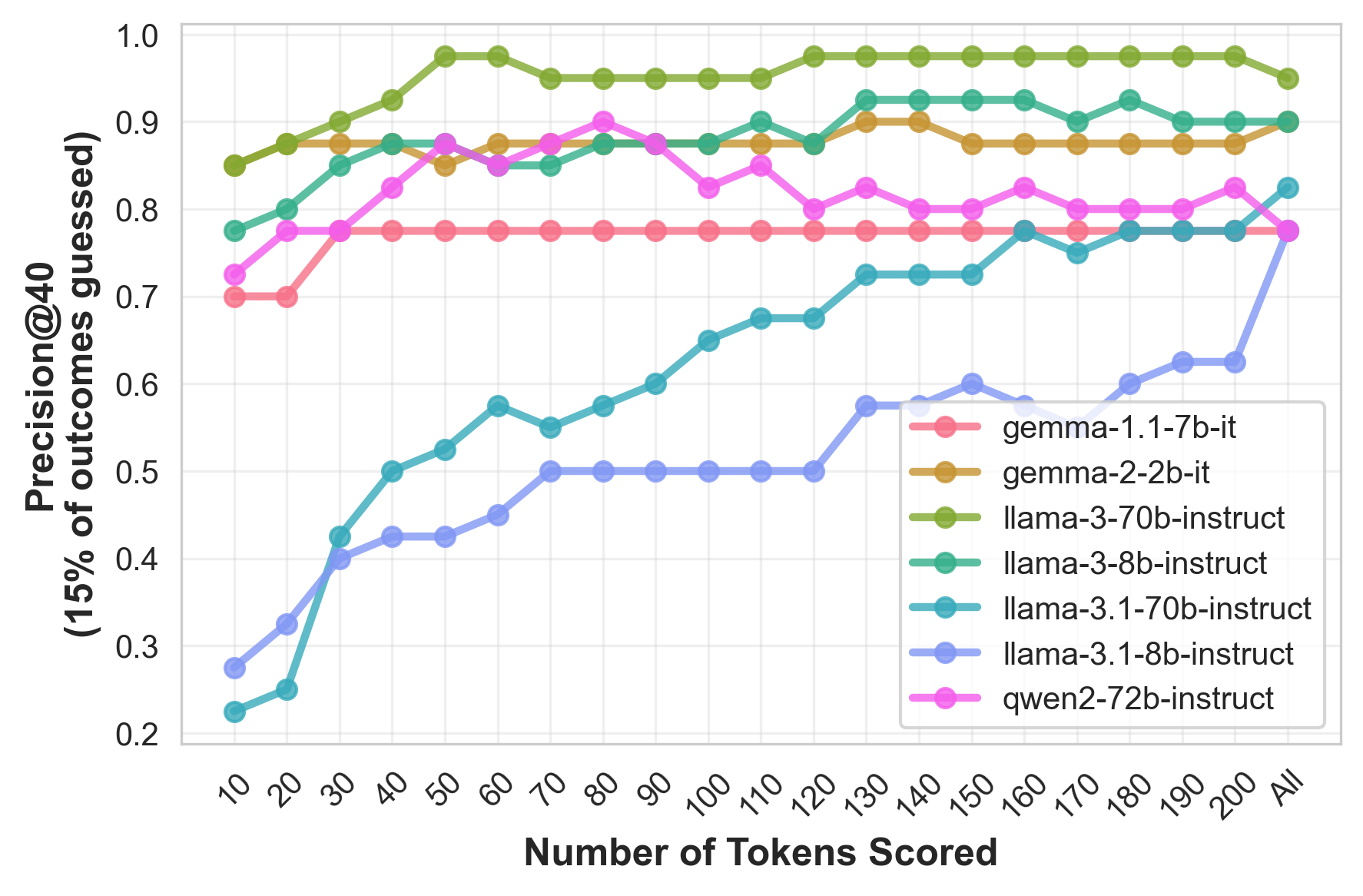}}
    \caption{
Precision@30 (15\% of outputs) as a function of tokens scored for open-weight models. Many models reach near-optimal precision with as few as 30 tokens, suggesting substantial efficiency gains are possible by truncating log-probability calculations early.
    }
    \label{precision_vs_tokens_scored}
  \end{center}
\end{figure}

For open-weight models, we compute the average token-level log-probability of the observed response under each candidate model using teacher forcing, i.e., the log-probability assigned to each answer token conditioned on the prompt and all preceding answer tokens.
For proprietary models, we generate a single response per prompt and compute character-level similarity to the observed response.

Figure~\ref{live_model_identification} shows that open-weight models can be identified with high precision ($>80\%$, except \texttt{qwen2-72b-instruct}, which has a slightly lower precision).
Proprietary models achieve lower but above chance precision (approximately $40-50\%$).

The outputs contain 315 tokens, on average.
Figure~\ref{precision_vs_tokens_scored} demonstrates that for open-weight models, we do not need to score all generated tokens to achieve high precision. Scoring only the first few tokens could reduce the computational cost of real-world model identification.

\section{Factors Influencing Ease of Movement}
\label{app:regressions}
We conducted regressions on the best $\Delta$ rank achieved from strategic voting, prompting, or nomination. We used ridge regression with the following factors for each model $A$: The factors we consider are $A$’s variance in win rate against different models, $A$’s starting rank and Elo, the percent of battles in which $A$ tied, the number of models in $A$’s model family, the gap in Elo between $A$ and the model one rank higher, $A$'s win rate and total battles against the model one rank higher, the number of battles in $A$'s best language for strategic prompting, the number of total battles involving $A$, and the ``Elo density'' (defined as the number of models with Elo 0-50 points higher than the current model).

\clearpage

\begin{table}
\centering
\caption{Regression coefficients for the best $\Delta$ rank achieved from strategic voting (RMSE=2.75).}
\begin{tabular}{lr}
\hline
Feature & Coefficient \\
\hline
elo density vs above              &  0.354919 \\
winrate variance               &  0.079747 \\
\# total battles                & -0.064431 \\
elo gap vs model above                  &  0.028544 \\
winrate vs model above                    & -0.027852 \\
model Elo                      &  0.026598 \\
\# battles vs model above             & -0.021019 \\
tie rate                       & -0.015949 \\
\# battles in best prompt lang  & -0.011382 \\
model rank                     & -0.009996 \\
family size                    & -0.009464 \\
\hline
\end{tabular}

\vspace{0.1em}
\begin{tabular}{lr}
Train $R^2$ & 0.660 \\
\end{tabular}
\end{table}

\begin{table}
\centering
\caption{Regression coefficients for the best $\Delta$ rank achieved from strategic prompting (RMSE=2.67).}
\begin{tabular}{lr}
\hline
Feature & Coefficient \\
\hline
\# battles in best prompt lang & -0.337490 \\
elo gap vs model above                  & -0.126945 \\
elo density vs above              &  0.111206 \\
winrate variance                    &  0.069611 \\
family size                    & -0.066190 \\
winrate vs model above                     &  0.040990 \\
\# battles vs model above             & -0.039253 \\
tie rate                       &  0.036262 \\
model rank                     &  0.027060 \\
\# total battles                & -0.016139 \\
model Elo                      & -0.012417 \\
\hline
\end{tabular}

\vspace{0.1em}
\begin{tabular}{lr}
Train $R^2$ & 0.518 \\
\end{tabular}
\end{table}

\begin{table}[h]
\centering
\caption{Regression coefficients for the best $\Delta$ rank achieved from strategic nomination (RMSE=1.85).}
\begin{tabular}{lr}
\hline
Feature & Coefficient \\
\hline
elo density vs above       &    0.231206 \\
elo gap vs model above                  &  -0.200973 \\
winrate variance                     &  0.179882 \\
model Elo                       & 0.157270 \\
\# battles in best prompt lang  &  0.140642 \\
tie rate                       &  0.089433 \\
model rank                     & -0.077523 \\
\# battles vs model above             &  0.073610 \\
\# total battles                & -0.036676 \\
family size                    &  0.017748 \\
winrate vs model above                    &  0.001865 \\
\hline
\end{tabular}

\vspace{0.1em}
\begin{tabular}{lr}
Train $R^2$ & 0.557 \\
\end{tabular}
\end{table}

\onecolumn
\section{Scaling Experiments}\label{app:scaling}
\vspace{0.3cm}

We conducted experiments varying the number of votes and found that when varying the total dataset size from 25\%-200\% of the current dataset size, manipulating the same percentage of total votes gives a near-uniform rank increase. This means that as the dataset size increases, the number of votes needed to achieve a similar boost increases linearly.

\begin{figure}[thb]
  \vskip 0.2in
  \begin{center}
    \centerline{\includegraphics[width=0.6\columnwidth]{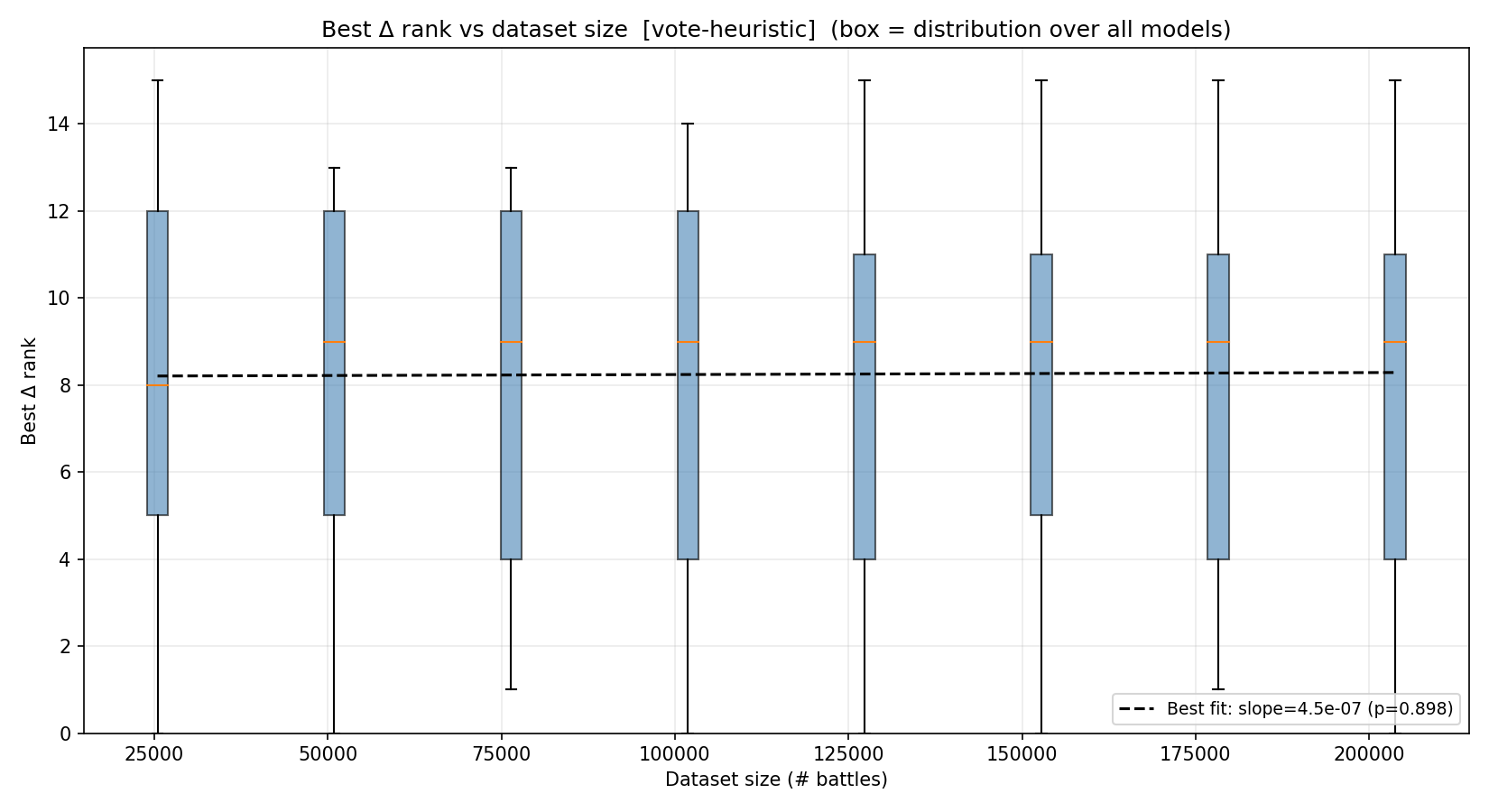}}
    \caption{
        Best delta rank achieved when manipulating a constant 10\% of battles on datasets of different sizes, on the \textit{heuristic voting} strategy. When plotting dataset size (x) against best rank achieved with $\leq$ 10\% battles modified (y), the slope is near-constant with p $>>$ 0.05 (not significantly different from a line with slope 0). This indicates that the number of manipulations needed to achieve similar rank boosts scales linearly with dataset size.
    }
    \label{fig:scaling_heuristic_voting}
  \end{center}
\end{figure}

We manipulated a constant $\sim$ 10\% of battles as the dataset size increases. For experiments over 100K examples, we resampled battles from the existing 100K dataset. Then, we plotted dataset size vs. maximum rank boost when manipulating that constant proportion of votes. For heuristic voting, greedy voting, and combined prompting + nomination, the slope is near-constant with $p >> 0.05$ 
(that is, not significantly different from a line with slope 0). This indicates that the number of manipulations needed to achieve similar rank boosts scales linearly with dataset size. For prompting and nominations, the slope is also near-constant, though we do find that manipulations become slightly easier as dataset size increases (though these attacks depend on the proportion of models or prompts to total votes, so this result may vary by leaderboard). Figures \ref{fig:scaling_heuristic_voting}-\ref{fig:scaling_combo} show the effects of scaling dataset size on the number of manipulations needed.  Across experiments, the number of manipulations scales linearly or close to linearly.

\begin{figure}[thb]
  \vskip 0.2in
  \begin{center}
    \centerline{\includegraphics[width=0.5\columnwidth]{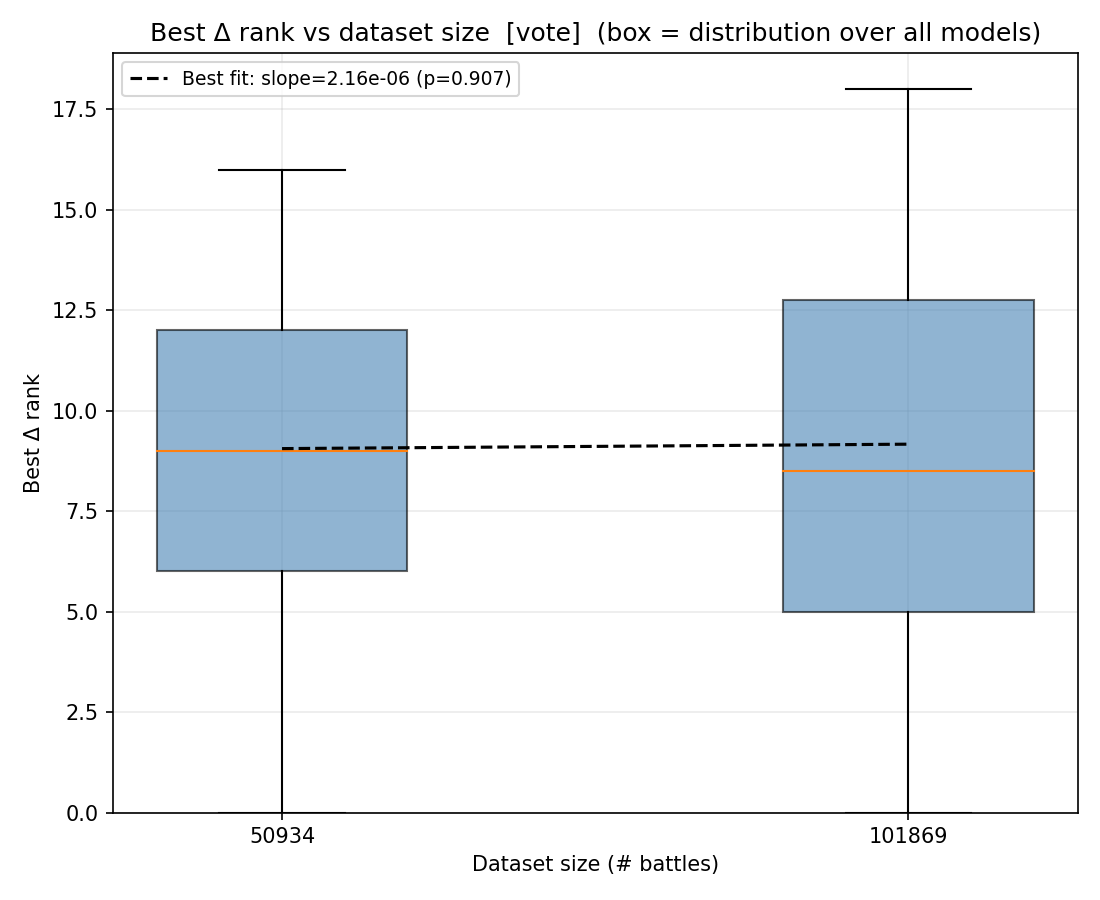}}
    \caption{
        Best delta rank achieved when manipulating a constant 10\% of battles on datasets of different sizes, on the \textit{greedy voting} strategy. When plotting dataset size (x) against best rank achieved with $\leq$ 10\% battles modified (y), the slope is near-constant with p $>>$ 0.05 (not significantly different from a line with slope 0). This indicates that the number of manipulations needed to achieve similar rank boosts scales linearly with dataset size.
    }
    \label{precision_vs_tokens_scored}
  \end{center}
\end{figure}

\begin{figure}[thb]
  \vskip 0.2in
  \begin{center}
    \centerline{\includegraphics[width=0.6\columnwidth]{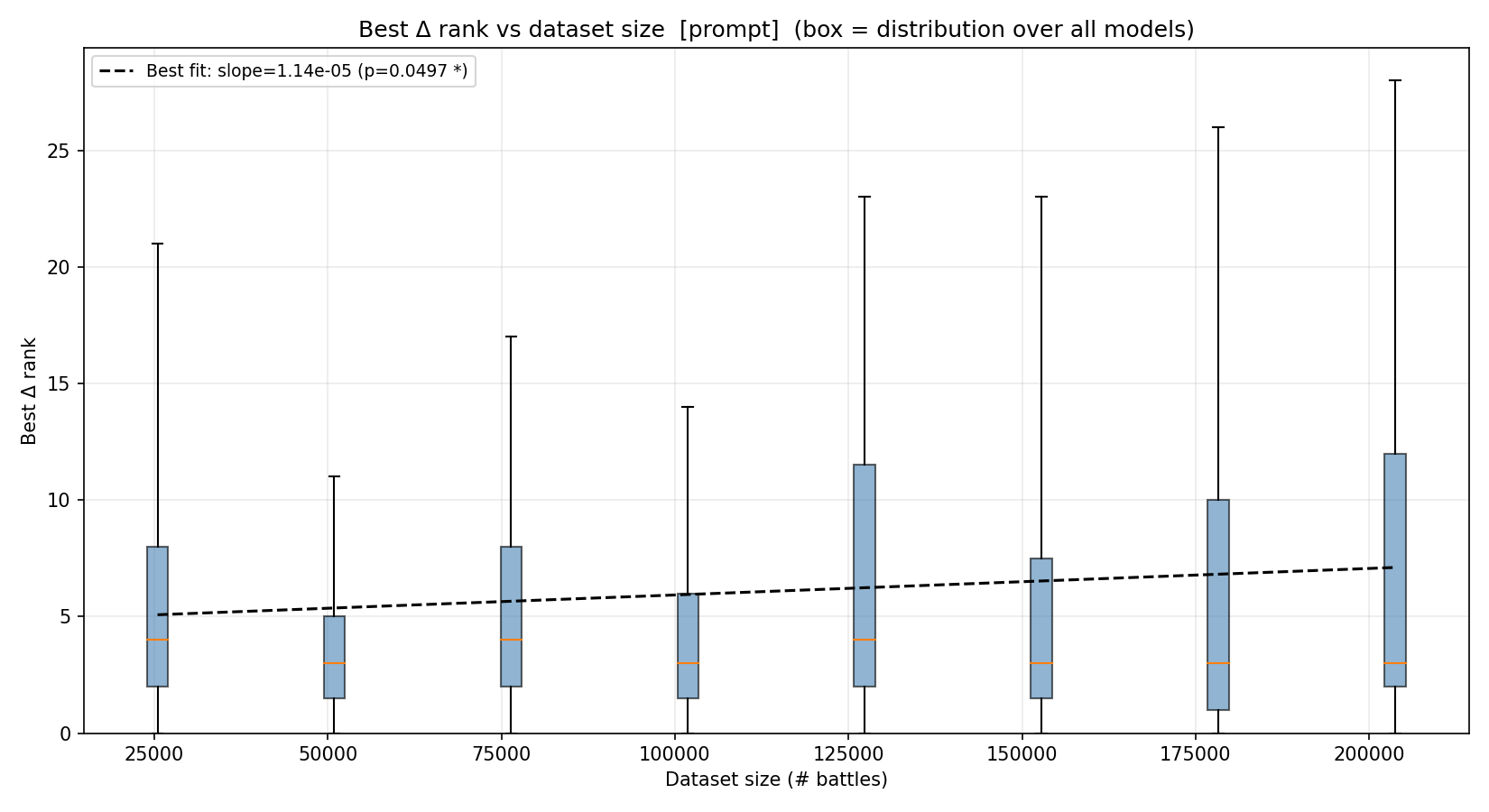}}
    \caption{
        Best delta rank achieved when manipulating a constant 10\% of battles on datasets of different sizes, with the \textit{strategic prompting} strategy. When plotting dataset size (x) against best rank achieved with $\leq$ 10\% battles modified (y), the slope is very slightly positive (p $<$ 0.05), meaning that the manipulation becomes slightly easier as dataset size increases.
    }
    \label{precision_vs_tokens_scored}
  \end{center}
\end{figure}

\begin{figure}[thb]
  \vskip 0.2in
  \begin{center}
    \centerline{\includegraphics[width=0.6\columnwidth]{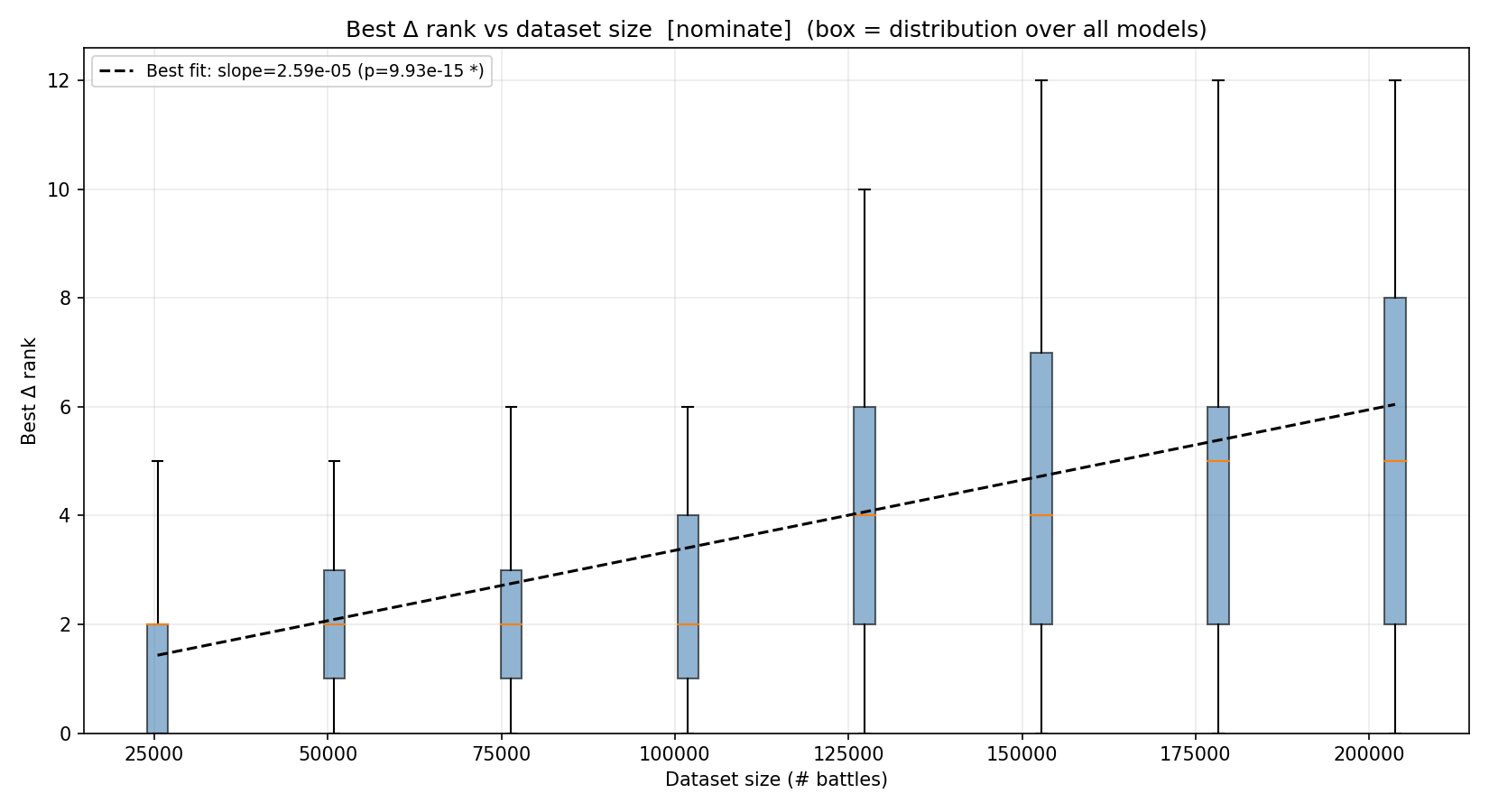}}
    \caption{
        Best delta rank achieved when manipulating a constant fraction of models on datasets of different sizes, using the \textit{strategic nomination} strategy. When plotting dataset size (x) against best rank achieved (y) with $\leq$ 10 model copies inserted (number of battles per model copy thus scales with dataset size), the slope is slightly positive (p $<$ 0.05), meaning the attack is somewhat easier as dataset size increases with the same proportion of models changed (though note that the proportion of models to total votes may vary by leaderboard).
    }
    \label{precision_vs_tokens_scored}
  \end{center}
\end{figure}

\begin{figure}[thb]
  \vskip 0.2in
  \begin{center}
    \centerline{\includegraphics[width=0.6\columnwidth]{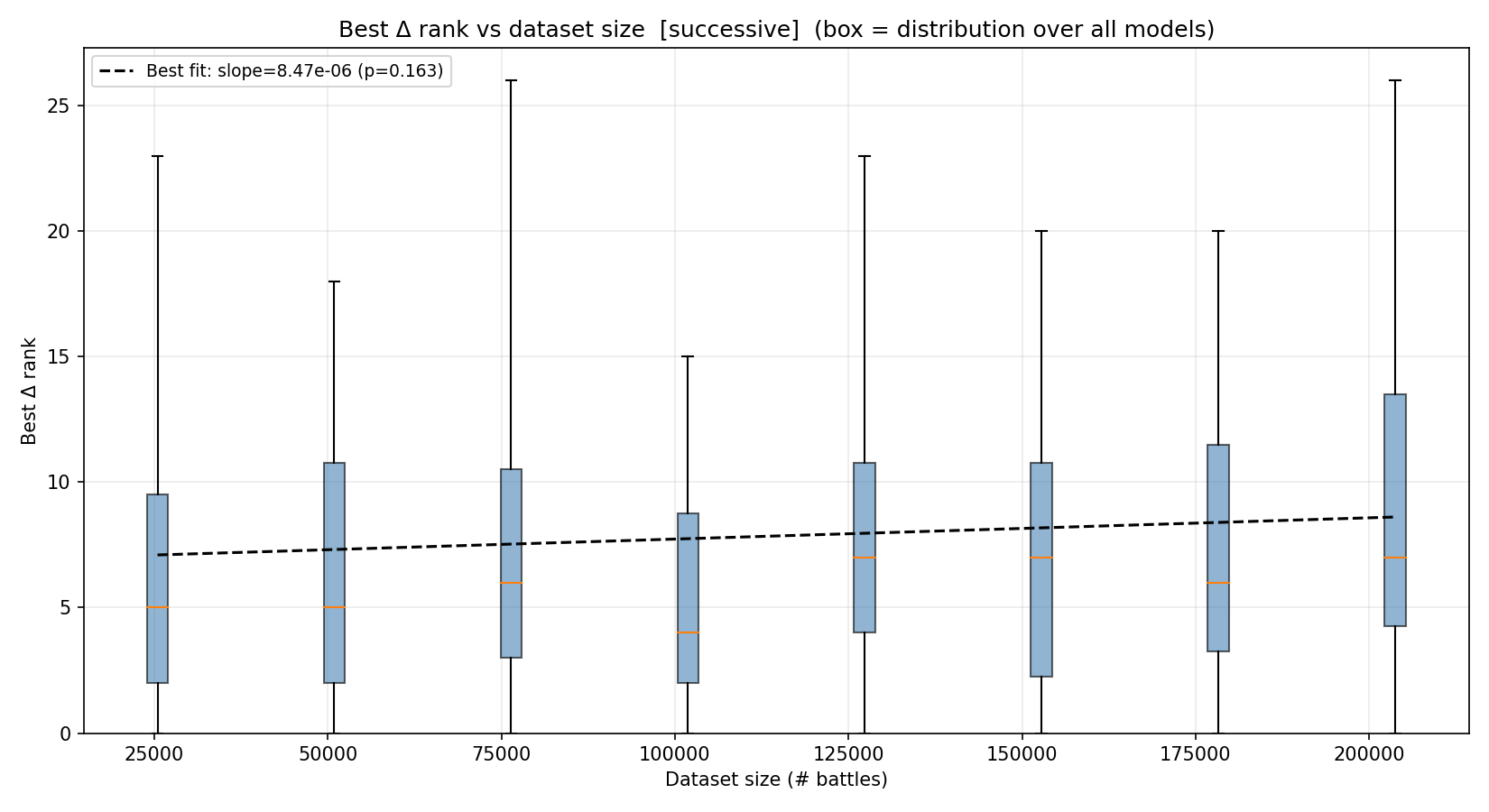}}
    \caption{
        Best delta rank achieved when manipulating a constant 10\% of battles on datasets of different sizes, when combining the strategic nomination and prompting strategies in succession. When plotting dataset size (x) against best rank achieved with $\leq$ 10\% battles modified (y), the slope is near-constant with p $>>$ 0.05 (not significantly different from a line with slope 0). This indicates that the number of manipulations needed to achieve similar rank boosts scales linearly with dataset size.
    }
    \label{fig:scaling_combo}
  \end{center}
\end{figure}

\section{Strategic Prompting: Changing Subcategories}\label{sec:prompting-other-subcats}

\begin{figure}[thb]
  \vskip 0.2in
  \begin{center}
    \centerline{\includegraphics[width=0.6\columnwidth]{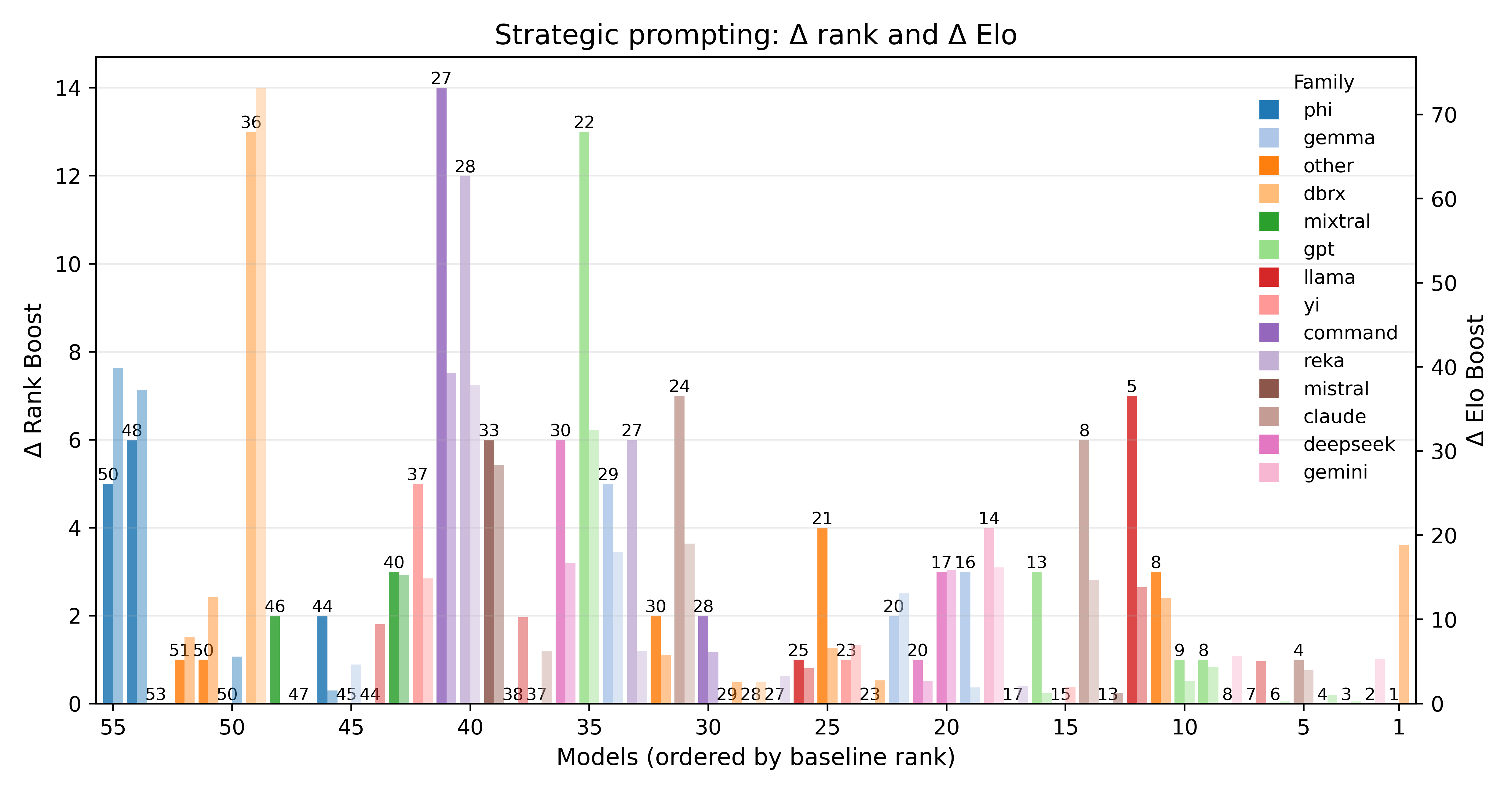}}
    \caption{
        Proof-of-concept that strategic prompting can use any definable, consistent subcategories: best delta rank and Elo achieved under strategic prompting, where prompt categories are based on whether the prompt is labeled as involving complexity, creativity, domain knowledge, and/or math. Rank boosts (up to 14 places) are comparable to those from strategic prompting using language for prompt subcategories.
    }
    \label{fig:prompt_by_category}
  \end{center}
\end{figure}

Here, we test strategic prompting with a different subcategory split as proof-of-concept that strategic prompting can use any definable, consistent subcategories.  We find rank boosts of up to 14 places, comparable to those from strategic prompting using language for prompt subcategories (Figure \ref{fig:prompt_by_category}).

\newpage
\section*{NeurIPS Paper Checklist}

\begin{enumerate}

\item {\bf Claims}
    \item[] Question: Do the main claims made in the abstract and introduction accurately reflect the paper's contributions and scope?
    \item[] Answer: \answerYes{} 
    \item[] Justification: Yes, the abstract and introduction state the claims made in the paper regarding leaderboard manipulation.

\item {\bf Limitations}
    \item[] Question: Does the paper discuss the limitations of the work performed by the authors?
    \item[] Answer: \answerYes{} 
    \item[] Justification: Yes, limitations are discussed in Appendix \ref{app:limitations}.

\item {\bf Theory assumptions and proofs}
    \item[] Question: For each theoretical result, does the paper provide the full set of assumptions and a complete (and correct) proof?
    \item[] Answer: \answerNA{} 
    \item[] Justification: We do not contribute any proofs.

    \item {\bf Experimental result reproducibility}
    \item[] Question: Does the paper fully disclose all the information needed to reproduce the main experimental results of the paper to the extent that it affects the main claims and/or conclusions of the paper (regardless of whether the code and data are provided or not)?
    \item[] Answer: \answerYes{} 
    \item[] Justification: Section \ref{sec:methods} provides a link to the (public) data we use and a detailed description of our methods, which are fairly straightforward to implement. We also plan to release code with the camera-ready.

\item {\bf Open access to data and code}
    \item[] Question: Does the paper provide open access to the data and code, with sufficient instructions to faithfully reproduce the main experimental results, as described in supplemental material?
    \item[] Answer: \answerNo{} 
    \item[] Justification: The data is publicly available. We plan to release the code shortly.

\item {\bf Experimental setting/details}
    \item[] Question: Does the paper specify all the training and test details (e.g., data splits, hyperparameters, how they were chosen, type of optimizer) necessary to understand the results?
    \item[] Answer: \answerYes{} 
    \item[] Justification: We specify the details of our experimental methodology; none of the experiments involved training new models, so there are no training-related hyperparameters to report.

\item {\bf Experiment statistical significance}
    \item[] Question: Does the paper report error bars suitably and correctly defined or other appropriate information about the statistical significance of the experiments?
    \item[] Answer: \answerYes{} 
    \item[] Justification: We report $p$-values for the scaling experiments in Appendix \ref{app:scaling} and RMSE for the regressions in Appendix \ref{app:regressions}. For Figures 1-4, the main plots already show the full distribution of attack effect sizes (i.e., the range of observed boosting effects) over all tested models.

\item {\bf Experiments compute resources}
    \item[] Question: For each experiment, does the paper provide sufficient information on the computer resources (type of compute workers, memory, time of execution) needed to reproduce the experiments?
    \item[] Answer: \answerYes{} 
    \item[] Justification: Yes, we describe computational feasibility in section \ref{sec:feasibility}.
    
\item {\bf Code of ethics}
    \item[] Question: Does the research conducted in the paper conform, in every respect, with the NeurIPS Code of Ethics \url{https://neurips.cc/public/EthicsGuidelines}?
    \item[] Answer: \answerYes{} 
    \item[] Justification: Yes, we have read and adhere to the code of ethics. Defenses to mitigate security risks from leaderboard manipulation research are discussed in Section \ref{sec:defenses}.

\item {\bf Broader impacts}
    \item[] Question: Does the paper discuss both potential positive societal impacts and negative societal impacts of the work performed?
    \item[] Answer: \answerYes{} 
    \item[] Justification: Yes, we discuss impacts in the conclusion and impact statement.
    
\item {\bf Safeguards}
    \item[] Question: Does the paper describe safeguards that have been put in place for responsible release of data or models that have a high risk for misuse (e.g., pre-trained language models, image generators, or scraped datasets)?
    \item[] Answer: \answerNA{} 
    \item[] Justification: We do not release data or models.

\item {\bf Licenses for existing assets}
    \item[] Question: Are the creators or original owners of assets (e.g., code, data, models), used in the paper, properly credited and are the license and terms of use explicitly mentioned and properly respected?
    \item[] Answer: \answerYes{} 
    \item[] Justification: Yes, we provide links to the LMArena data used and note its CC-BY-4.0 license in footnote \ref{data}.

\item {\bf New assets}
    \item[] Question: Are new assets introduced in the paper well documented and is the documentation provided alongside the assets?
    \item[] Answer: \answerNA{} 
    \item[] Justification: We do not introduce new assets.

\item {\bf Crowdsourcing and research with human subjects}
    \item[] Question: For crowdsourcing experiments and research with human subjects, does the paper include the full text of instructions given to participants and screenshots, if applicable, as well as details about compensation (if any)? 
    \item[] Answer: \answerNA{} 
    \item[] Justification: No human subjects research was performed.

\item {\bf Institutional review board (IRB) approvals or equivalent for research with human subjects}
    \item[] Question: Does the paper describe potential risks incurred by study participants, whether such risks were disclosed to the subjects, and whether Institutional Review Board (IRB) approvals (or an equivalent approval/review based on the requirements of your country or institution) were obtained?
    \item[] Answer: \answerNA{} 
    \item[] Justification: No human subjects research was performed.

\item {\bf Declaration of LLM usage}
    \item[] Question: Does the paper describe the usage of LLMs if it is an important, original, or non-standard component of the core methods in this research? Note that if the LLM is used only for writing, editing, or formatting purposes and does \emph{not} impact the core methodology, scientific rigor, or originality of the research, declaration is not required.
    \item[] Answer: \answerNA{} 
    \item[] Justification: The core methods do not involve LLMs as an important/original/non-standard component.

\end{enumerate}

\end{document}